\documentclass{article}

\PassOptionsToPackage{numbers, compress}{natbib}

\usepackage[preprint]{neurips_2026}

\usepackage[utf8]{inputenc}
\usepackage[T1]{fontenc}
\usepackage[colorlinks=true,linkcolor={blue!70!black},citecolor={blue!70!black},urlcolor={blue!70!black}]{hyperref}
\usepackage{url}
\usepackage{booktabs}
\usepackage{amsfonts}
\usepackage{microtype}
\usepackage{xcolor}
\usepackage{graphicx}
\usepackage{multirow}
\usepackage{cleveref}
\usepackage{caption}
\usepackage{xcolor}
\usepackage{tikz}
\definecolor{gamecolor1}{RGB}{253, 230, 153}
\definecolor{gamecolor2}{RGB}{235, 154, 155}
\definecolor{gamecolor3}{RGB}{180, 216, 170}
\definecolor{gamecolor4}{RGB}{166, 193, 228}
\definecolor{gamecolor5}{RGB}{245, 179, 108}
\definecolor{gamecolor6}{RGB}{180, 168, 210}
\DeclareRobustCommand{\gamebadge}[2]{\tikz[baseline=(t.base)]{\node[fill=#1,rounded corners=2pt,inner sep=1.5pt](t){\textsc{#2}};}}
\title{Reason to Play: Behavioral and Brain Alignment \\ Between Frontier LRMs and Human Game Learners}

\author{
  Botos Csaba\textsuperscript{1}\thanks{Equally contributing first authors} \quad
  Sreejan Kumar\textsuperscript{2,3}\footnotemark[1] \\
  Austin Tudor David Andrews\textsuperscript{1} \quad
  Laurence Hunt\textsuperscript{1} \quad
  Chris Summerfield\textsuperscript{1} \quad
  Joshua B. Tenenbaum\textsuperscript{4} \\
  Rui Ponte Costa\textsuperscript{1}\thanks{Equally contributing senior authors} \quad
  Marcelo G. Mattar\textsuperscript{3}\footnotemark[2] \quad
  Momchil Tomov\textsuperscript{5}\footnotemark[2] \\[6pt]
  \textsuperscript{1}University of Oxford \quad
  \textsuperscript{2}Columbia University \quad
  \textsuperscript{3}New York University \\
  \textsuperscript{4}Massachusetts Institute of Technology \quad
  \textsuperscript{5}Harvard University
}

\begin{document}

\maketitle


\begin{abstract}
Humans rapidly learn abstract knowledge when encountering novel environments and flexibly deploy this knowledge to guide efficient and intelligent action. Can modern AI systems learn and plan in a similar way? We study this question using a dataset of complex human gameplay with concurrent fMRI recordings, in which participants learn novel video games that require rule discovery, hypothesis revision, and multi-step planning. We jointly evaluate models by their ability to play the games, match human learning behavior, and predict brain activity during the same task, comparing a suite of frontier Large Reasoning Models (LRMs) against model-free and model-based deep reinforcement learning agents and a Bayesian theory-based agent. We find that frontier LRMs most closely match human behavioral patterns during game discovery and predict brain activity an order of magnitude better than both reinforcement learning alternatives across cortical and subcortical regions, with effects robust to permutation controls. Through targeted manipulations, we further show that brain alignment reflects the model's in-context representation of the game state rather than its downstream planning or reasoning.
Our results establish LRMs as compelling computational accounts of human learning and decision making in complex, naturalistic environments.
Project page with interactive replays: \url{https://botcs.github.io/reason-to-play/}
\end{abstract}

\section{Introduction}
\label{sec:intro}

Humans rapidly learn the abstract structure of unfamiliar environments (e.g., a novel tool, a new social situation) and flexibly deploy this knowledge to guide intelligent action \citep{tenenbaum2011grow}. This capacity for abstract rule induction under minimal priors is among the most distinctive features of human cognition, and games have become a powerful sandbox for studying it \citep{allen2024using}. \citet{schaul2013video} designed the Video Game Description Language (VGDL), a framework that specifies games
compositionally through object types, interaction rules, and win/loss
conditions. VGDL games are uniquely suited for studying complex human reasoning behavior, occupying a niche that other benchmarks do not: unlike static reasoning tasks, which elicit a single chain of thought per problem, VGDL is interactive, with every step offering a new opportunity for hypothesis formation, verification, and revision; and unlike Atari-style benchmarks, where learning is dominated by sensorimotor pattern-matching over thousands of frames, VGDL confronts the agent with the harder problem of inferring object identities, interaction rules, and goal structure from interaction alone \citep{tsividis2021human}. The result is a dense, temporally structured record of human reasoning under uncertainty.

Video games have been equally central to artificial intelligence. The Deep RL
era began with DQN achieving human-level performance on Atari
\citep{mnih2015human}, and games have remained a guiding benchmark as the
field moved toward model-based agents that learn internal world models to
support planning \citep{ha2018world, hafner2023mastering,
wang2024efficientzero}. More recently, Large Language Models (LLMs) have
become the dominant paradigm \citep{achiam2023gpt}, despite persistent
criticism of their limited multi-step reasoning and planning capabilities
\citep{valmeekam2023planning}. Large Reasoning Models (LRMs) respond by
generating explicit chains of thought 
\citep{liu2024deepseek, deepseekai2025deepseekv32, qwen2026qwen35}.
Can these models learn and plan like humans? Specifically, can LRMs learn \emph{interactively} by rapidly extracting generalizable knowledge from dynamic environments and flexibly deploying it to guide action?

\begin{figure}[t]
    \centering
  \includegraphics[width=\linewidth]{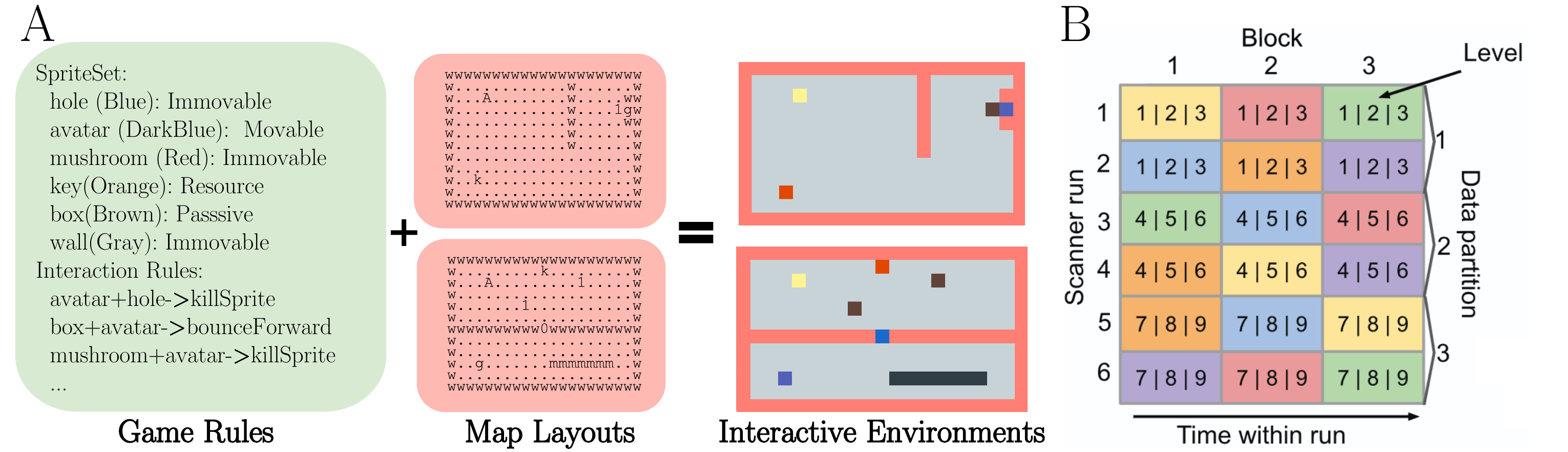}
    \caption{\textbf{VGDL game paradigm.} \textbf{(A)} Games are defined by combining game rules with map layouts to produce interactive environments.  \textbf{(B)} Example Trial Structure of VGDL-fMRI Dataset. Color denotes game names: (\gamebadge{gamecolor1}{Bait}, \gamebadge{gamecolor2}{Chase}, \gamebadge{gamecolor3}{Helper}, \gamebadge{gamecolor4}{Lemmings}, \gamebadge{gamecolor5}{Plaque Attack}, \gamebadge{gamecolor6}{Zelda}). All participants played the same level progression structure with randomized game order. The subsequent levels reveal new rules incrementally. The Interactive Catalogue~\ref{app:interactive_supp} lets readers try each game in the browser and browse all participant and LRM agent gameplay replays. Project page: \url{https://botcs.github.io/reason-to-play/}
  }
  \vspace{-2.8mm}
\label{fig:overview}
\end{figure}

Answering this question requires more than behavioral benchmarks alone. Existing evaluations of general LLM world-modeling \citep{hendriksen2025adapting,yang2024evaluating} and interactive agentic game-like benchmarks like ARC-AGI-3 \citep{foundation2026arc} and AI GameStore \citep{ying2026ai} test what models \emph{do}, but reveal little about what their internal representations \emph{look like} while doing it. Mechanistic interpretability has emerged as a major framework for characterizing such internal computations \citep{olah2020zoom,elhage2022toy}, but it lacks reference to representations from other intelligent systems performing the same task. Human brain activity offers a complementary external signal: when paired with a matched interactive task, it offers a ground-truth target for
the internal representations humans use to perform that task, which are not directly observable from behavior alone (see \citet{mineault2026cognitive}). The VGDL-fMRI paradigm of \citet{tomov2023neural} uniquely combines human VGDL trajectories and concurrent brain activity during the same game-learning process, providing a unique target for both human behavior and representations.

A parallel gap exists in computational neuroscience, where the rise of deep learning has redefined what counts as a model of the brain. Task-optimized Deep Neural Networks (DNNs) for language, vision, and audition are now the best predictors of brain activity in their corresponding sensory cortical regions \citep{yamins2016using,schrimpf2021neural,kell2018task}. But these results have focused largely on sensory perception, rather than interactive cognition, in which an agent must cycle between perception, learning, and action. Initial work in the interactive domain has found that deep learning models trained from scratch on each game produce brittle encoding signals that approach random-initialization baselines \citep{paugam2025brittle}. Frontier LRMs offer a path past this limitation: unlike previously used task-optimized DNNs, they can learn new tasks without per-task training, and may therefore carry the rich, transferable representations that brittle from-scratch models lack.

Our contributions are as follows.
\textbf{(i)} We provide the first systematic evaluation of frontier LRMs on human-like learning of interactive video games, testing eight models (Qwen~3.5
9B/27B/35B-a3B/122B-a10B/397B-a17B \citep{qwen2026qwen35}; DeepSeek V3.2/V4-Flash/V4-Pro
\citep{deepseekai2025deepseekv32, deepseekai2026deepseekv4}) across all
twelve VGDL-fMRI games and thirty-two human participants. Earlier work applying LLMs to VGDL reported limited success \citep{colas2026language}, but those evaluations used single-step prompting and pre-reasoning-era LLMs (\emph{LLaMA-3.1-70B}); no subsequent work has systematically evaluated current frontier reasoning models with multi-step dialogue on this task.
\textbf{(ii)} On behavior, frontier LRMs match human game-learning
trajectories on both learning efficiency and capability, whereas deep RL baselines
require orders of magnitude more experience to learn the games while being less capable.
%
\textbf{(iii)} On brain encoding, LRM representations predict BOLD activity
across multiple regions of interest an order of magnitude better than DDQN,
EfficientZero, and HRR baselines, and $6.4$--$6.9\times$ above
same-architecture random-initialization controls (paired $t > 14$,
$p < 10^{-11}$; $n=21$)
\textbf{(iv)} These results are achieved off-the-shelf, without fine-tuning,
a privileged ontology, or per-game training. We release the evaluation harness, all replay traces, extracted hidden-state features, and the full analysis pipeline alongside the paper. Across the eight models, twelve games, and two rationale modes in our grid, this yields over 100,000 reasoning traces---a rich dataset for the dynamics of in-context problem solving.





\section{Methods}
\label{sec:methods}
\subsection{Dataset}

\label{sec:dataset}
The VGDL-fMRI dataset comprises data from 32 healthy adult participants (17 male and 15 female; 19–36 years; all right-handed) who learned to play VGDL games while undergoing fMRI. Games were grid-worlds spanning a heterogeneous space of mechanics, including pushing and obstruction, agent-tracking pursuit, multi-step subgoaling, hazard avoidance, and cooperative guidance, requiring participants to infer qualitatively different rule structures across games. Each participant played six games over six scanner runs, with each run divided into three blocks of three levels each, for a total of nine levels per game (Figure~\ref{fig:overview}B). Each level was played for 60 seconds total, with episodes restarting immediately on the same level after a win or loss; once the 60 seconds elapsed, the next level began regardless of performance. Levels were explicitly designed to sustain learning: later levels vary layouts and introduce interaction opportunities not present earlier.

\subsection{Baseline model classes}
\label{sec:baselines}
\paragraph{Model-free deep RL}
We use a Double DQN (DDQN) \cite{van2016deep} as a model-free baseline, following \citet{tomov2023neural}, but with two modifications to better support representation analysis: (i) per-level training for a fixed budget of 100k gradient updates, capturing before/after-learning representations, and (ii) extensive per-game hyperparameter tuning (256 configurations $\times$ 4 Hyperband stages per game) rather than a single shared configuration. Full details are in Appendix~\ref{app:ddqn}.

\paragraph{Model-based deep RL}
We additionally evaluate EfficientZeroV2 (EZV2, \cite{wang2024efficientzero}), a model-based deep RL agent that learns a latent dynamics model to support planning via Monte Carlo tree search. Unlike DDQN, EfficientZero learns an internal transition model of the environment, making it a model-based deep RL counterpart to the model-free DDQN. Full details are in Appendeix~\ref{app:ez}.

\paragraph{Bayesian theory-based RL}
The Explore–Model–Plan Agent (EMPA) \cite{tsividis2021human} is a model-based symbolic baseline that explicitly represents hypothesized causal rules governing object interactions. EMPA approximates Bayesian inference over the space of all possible VGDL rules to infer a running posterior of the rules (the \textit{theory}) of the current game. Planning proceeds by simulating the outcomes of candidate action sequences using a VGDL simulator grounded in the inferred theory.

\subsection{Large reasoning models}
\begin{figure}[t]
\centering
\includegraphics[width=0.65\textwidth]{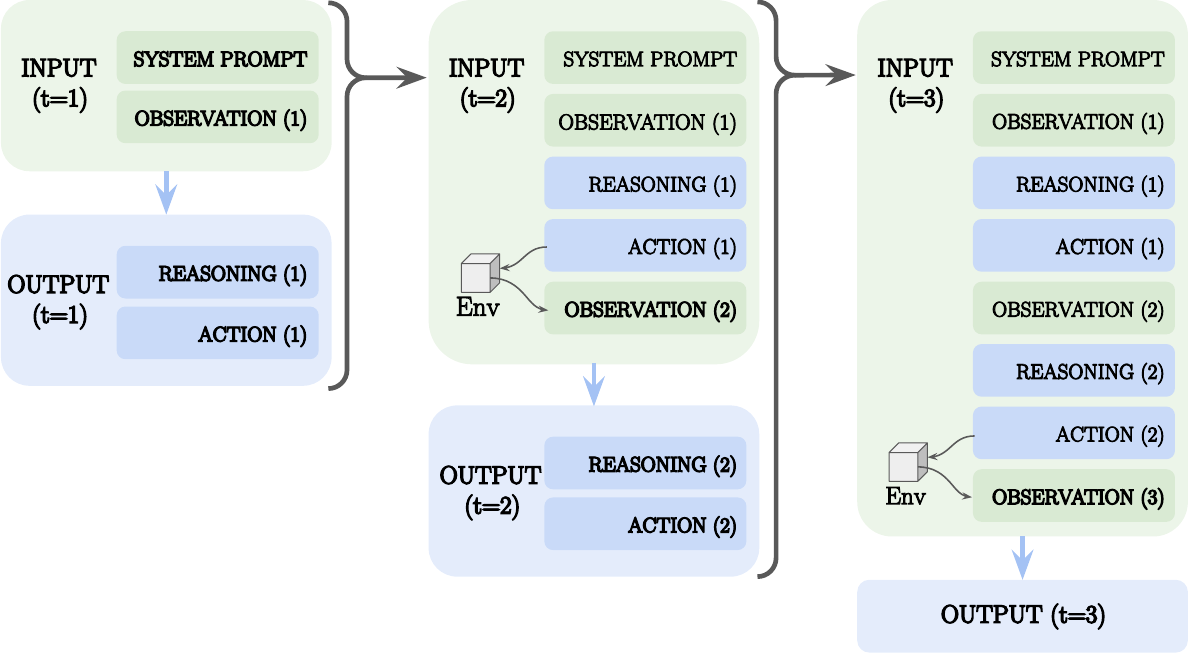}
\caption{\textbf{Multi-turn dialogue format for generating gameplay.} At each step, the model receives the current observation as a user message and responds with an action. In the \emph{copied-reasoning} condition, the model's hidden reasoning trace is copied into the context as the stated rationale on the next turn. The full conversation history accumulates in the context window, giving the model access to all past observations, actions, and (optionally) rationales, without any external memory mechanism.}
\label{fig:dialouge}
\end{figure}
\label{sec:lrms}
We evaluate eight LRMs, a class of LLMs post-trained to produce explicit reasoning traces before committing to an output. This training is typically done via RL on verifiable reasoning tasks such as mathematics and code \citep{liu2024deepseek}. Unlike the DDQN and EfficientZero, LRMs acquire no task-specific weights through gameplay; all task-relevant knowledge is constructed in-context from the observation stream and the system prompt (Figure~\ref{fig:dialouge}).

\paragraph{Models}
We test five models from the Qwen~3.5 family \citep{qwen2026qwen35} spanning dense and Mixture-of-Experts (MoE) architectures: \textbf{Qwen3.5-9B} (9B dense), \textbf{Qwen3.5-27B} (27B dense), \textbf{Qwen3.5-35B-a3B} (35B total, $\sim$3B active per token, MoE), \textbf{Qwen3.5-122B-a10B} (122B total, $\sim$10B active, MoE), and \textbf{Qwen3.5-397B-a17B} (397B total, $\sim$17B active, MoE). From the DeepSeek family, we test three models: \textbf{DeepSeek~V3.2} \citep{deepseekai2025deepseekv32}, a 685B-parameter MoE reasoning model; \textbf{DeepSeek~V4-Pro} and \textbf{DeepSeek~V4-Flash} \citep{deepseekai2026deepseekv4}. This set spans a wide range of scales, architectures, and post-training regimes while holding the evaluation protocol constant.

\paragraph{State abstraction and anonymization}
At each timestep the environment returns a structured VGDL state, which we transform into a text-based, symbolic observation using a deterministic formatter. To hide privileged semantics, internal engine identifiers (e.g., \texttt{goal1}, \texttt{key1}) are mapped to simple color identifiers (e.g. \texttt{GREEN}, \texttt{ORANGE}). The only object attributes revealed to the model are score, color, grid position, inventory and the outcome of the last trial upon a new trial---matching exactly the information available to human participants, who likewise see only color-anonymized sprites with no privileged semantic labels \citep{tomov2023neural}. Game-specific rules must be inferred from interaction.

\subsection{Multi-turn dialogue format}
\label{sec:multiturn}
\vspace{-2mm}

Unlike prior work that compressed the interaction into a single LRM call per step \citep{colas2026language}, we implement a multi-step dialogue paradigm (Figure~\ref{fig:dialouge}). Each game step maps to one conversational exchange: the \emph{user} message delivers the current observation (anonynmised game state) and the \emph{assistant} message contains the model's response.

\paragraph{Rationale modes} We test two conditions that differ only in what appears in the assistant message: \emph{Copied-reasoning.} The model runs in native reasoning mode. Its hidden chain-of-thought is copied into the assistant message as a \texttt{rationale} field alongside the chosen \texttt{action}. On subsequent turns the model can read its own prior reasoning, enabling multi-step hypothesis formation and revision.
\emph{Action-only.} The assistant message contains only the \texttt{action} field. The model's reasoning mode is disabled, thus no reasoning trace appears in context. This isolates the contribution of explicit reasoning and matches the input format used for brain encoding (\S\ref{sec:encoding}).
The user-message template (observation format, action log, feedback) is identical across both conditions; only the assistant-message schema differs.
\vspace{-1mm}
\paragraph{Suggestion levels} The system prompt encodes different amounts of prior knowledge about VGDL mechanics:
\emph{minimal} provides only the action space and response format;
\emph{elaborate} (headline condition) adds a general description of grid-world game mechanics and hints about rule discovery;
\emph{oracle} additionally provides the ground-truth game rules.
All behavioral results in the main text use the elaborate level; suggestion-level ablations are in the Supplementary Material.
\vspace{-1mm}
\paragraph{Level advancement} Unlike the original experiment, which advanced participants to the next level after a fixed 60-second window regardless of performance, we use a blocked curriculum: the agent must achieve two consecutive wins before advancing. This allows us to measure discovery (steps to first win) and execution (steps to subsequent wins) on each level, as well as to assess within-level learning. A global step budget caps each run. For the behavioral analysis, human behavior is forced to mimic the blocked curriculum by excluding episodes past two consecutive wins on each level and aborting the game if a level is not solved within the budget.

\section{Behavioral results}
\label{sec:behavior}
\vspace{-2.5mm}

\begin{figure}[t]
\centering
\includegraphics[width=\textwidth]{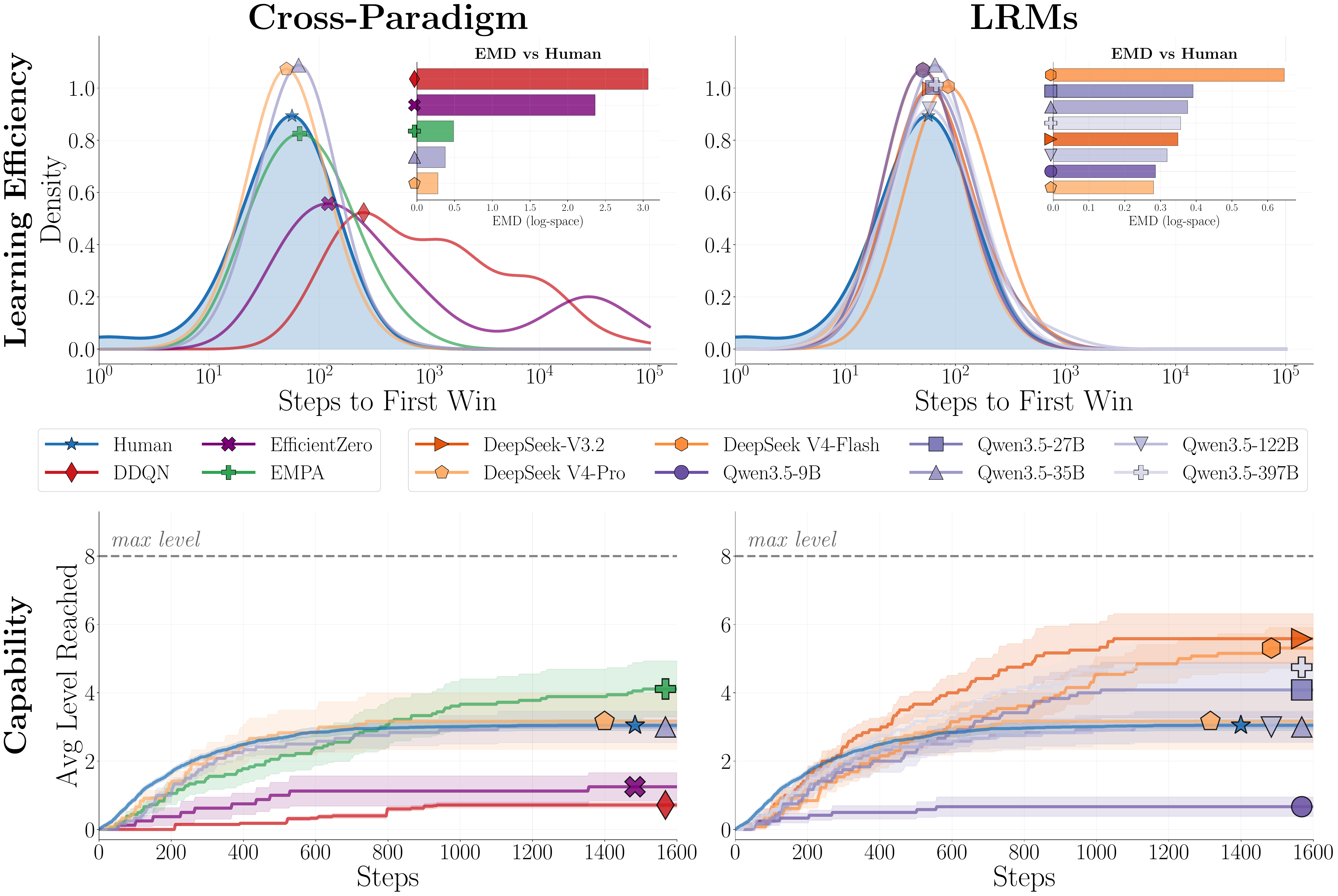}
  \caption{\textbf{Performance of LRM and baseline models.} Throughout the paper, cross-paradigm comparisons are anchored to two representative frontier LRMs: \textbf{DeepSeek V4-Pro}, the strongest \emph{behavioural} model (jointly best on the discovery KDE-EMD and on the level-progression curve in this figure), and \textbf{Qwen3.5-35B-A3B}, the strongest \emph{brain-encoding} model (\Cref{fig:encoding}B). The left column here and panel A of \Cref{fig:encoding} use this same pair. \textbf{(Top)} Learning efficiency: KDE of cumulative steps to first win, pooled across all games and levels (log scale). Human distribution shown as shaded fill; agents as curves with peak markers. Inset bars: log-space Earth Mover's Distance (EMD) between each agent's KDE and the human KDE in the same panel. \textbf{(Bottom)} Capability: average level reached as a function of cumulative steps under the blocked-curricula protocol (two consecutive wins required to advance; trajectories that stall hold their last level until the budget expires, pulling the average down). \textbf{(Left)} Three deep-RL/symbolic baselines plus the two anchor LRMs. \textbf{(Right)} The same data expanded to all eight evaluated LRMs (no baselines). Per-game breakdowns in \Cref{fig:discovery_cross_paradigm_supp,fig:discovery_llms_supp}.}
\label{fig:discovery_combined}
\end{figure}

We evaluate all agents on the twelve VGDL-fMRI games (six per cohort variant; five shared across both cohorts). Two headline measures quantify behavioural alignment with humans, both shown in \Cref{fig:discovery_combined}: \emph{discovery} -- cumulative steps to first win on a level (top row) -- captures how long an agent spends inferring the rules of an unseen level; \emph{capability progression} -- average level reached as a function of cumulative steps under the blocked-curriculum protocol (bottom row) -- captures how far through the 9-level sequence an agent advances under a fixed step budget. Two further measures, \emph{execution} (steps per subsequent win on the same level) and \emph{Kaplan--Meier survival} (fraction of levels solved over cumulative experience), are reported as supporting views in the Supplementary Material; we found them less informative about the behavioural traits we wanted to capture and keep them out of the main text. For LRMs, the headline condition is copied-reasoning with elaborate suggestion level (\S\ref{sec:multiturn}); an action-only condition that strips all reasoning from the conversation context is analysed separately (\S\ref{sec:action_only_supp}).
Human participants solved 75\% of level-instances (1303/1728), with substantial per-game variation ranging from 49\% on \textsc{Plaque Attack} to 93\% on \textsc{Zelda}. 

\subsection{LRMs match human discovery efficiency}
\label{sec:discovery}

We anchor cross-paradigm comparisons throughout the paper to two representative LRMs: \textbf{DeepSeek V4-Pro}, jointly strongest on discovery efficiency and on the capability progression curve (\Cref{fig:discovery_combined}), and \textbf{Qwen3.5-35B-A3B}, the strongest brain-encoding model (\Cref{fig:encoding}B). These two appear alongside the baselines in the left columns of both \Cref{fig:discovery_combined} and \Cref{fig:encoding}; the right-hand columns expand the comparison to the full LRM set.

\Cref{fig:discovery_combined} shows the distribution of discovery times (cumulative steps to first win per level) for all agents, pooled across games and levels. We quantify behavioral similarity using the log-space Earth Mover's Distance (EMD) between each agent's discovery distribution and the human reference, computed as the Wasserstein distance on log-transformed step counts to account for the three-order-of-magnitude range of discovery times.

The RL baselines occupy a regime far from human behavior: DDQN achieves a discovery EMD of $3.07$, and EfficientZero reaches $3.22$ once its game-specific warmup curriculum is counted as part of the steps spent before the first level-0 win. Both reflect the orders-of-magnitude more experience these agents need to fit their weights to solve the games. EMPA, which has access to a hand-coded ontology of game mechanics, achieves an EMD of $0.49$---substantially closer to human performance, but still worse than every LRM except DeepSeek V4-Flash.

Frontier LRMs cluster tightly around the human distribution. The most human-like individual model, DeepSeek V4-Pro, achieves a discovery EMD of $0.28$; the least human-like, DeepSeek V4-Flash, reaches $0.65$. V4-Pro is also the closest LRM to the human curve on the capability progression axis (\Cref{fig:discovery_combined}, bottom row), which is why we use it as our behavioural anchor (\Cref{tab:capability}). Across the LRM family this represents a $5$--$11\times$ reduction in discovery EMD relative to the deep-RL baselines. Per-game discovery distributions show that this aggregate pattern holds across individual games, although with heterogeneity across games: \textsc{Zelda} is nearly universally solved, while \textsc{Avoid George} and \textsc{Plaque Attack} remain challenging for all agents (\Cref{fig:discovery_cross_paradigm_supp,fig:discovery_llms_supp}).

\subsection{Discovery--execution gap in LRMs}
\label{sec:execution}

Per-game execution distributions for all agents are shown in \Cref{fig:execution_cross_paradigm_supp,fig:execution_llms_supp}; here we surface a single LRM-specific finding that we judge worth reporting in the main text.

If an agent has genuinely understood a level on its first win, subsequent attempts on the same level should be substantially shorter -- the rules are now known, only the rote execution remains. Humans show this strongly: the median number of steps per subsequent win is 32, well below the human discovery median. EMPA (median 34 steps, execution EMD $0.29$ against humans) and EfficientZero (median 40 steps, EMD $0.42$) re-execute comparably compactly once they have learned the rules; DDQN remains slow even in execution (median 189 steps, EMD $2.05$).

LRMs do not consolidate as compactly. Although their discovery distributions closely match human behaviour (\S\ref{sec:discovery}), their median execution times are $45$--$88$ steps across models and execution EMDs are $0.39$--$0.96$. Inspecting the reasoning traces (Appendix~\ref{app:interactive_supp}) shows that this is dominated by \emph{perseveration}: on a second attempt the model replays the exact trajectory of the winning first attempt, even when shorter routes have been articulated in its own rationale (e.g.\ door-key-door instead of just key-door). On stochastic levels, the smaller LRMs additionally fail to react to NPCs whose positions differ from the first run. This perseveration fades with scale across the Qwen3.5 family and is essentially absent in the DeepSeek models.

\subsection{Capability: solve rates across games}
\label{sec:capability}

Beyond timing, solve rate measures whether an agent can learn each game at all (\Cref{tab:capability}, \Cref{fig:trial_grid_vgfmri3_humans,fig:trial_grid_vgfmri4_humans,fig:trial_grid_vgfmri3_blocked,fig:trial_grid_vgfmri4_blocked}). The eight LRMs in the copied-reasoning condition solve between 11\% (Qwen3.5-9B, 12/108 level-instances) and 65\% (DeepSeek~V3.2, 70/108) of level-instances. The top three models---DeepSeek~V3.2 (65\%), V4-Flash (59\%), and Qwen3.5-397B (55\%)---approach human-level capability on several individual games (e.g., \textsc{Bait}, \textsc{Zelda}) while remaining far below humans on others (e.g., \textsc{Lemmings}, \textsc{Avoid George}; \Cref{fig:curriculum_llms_supp,fig:km_llms_supp}). DDQN solves 28\% (225/792) and EMPA solves 64\% (104/162) of their respective level-instances (\Cref{fig:curriculum_cross_paradigm_supp,fig:km_cross_paradigm_supp}).

The per-game pattern reveals that games requiring multi-step subgoaling and cooperative coordination (\textsc{Helper}, \textsc{Lemmings}) are harder for all agents, while games with simpler collect-and-avoid mechanics (\textsc{Bait}, \textsc{Zelda}) are more uniformly solved. Level progression curves (\Cref{fig:discovery_combined,fig:curriculum_cross_paradigm_supp,fig:curriculum_llms_supp}) show that 
LRMs progress through levels at a rate comparable to humans, while RL baselines plateau early.

\subsection{Qualitative analysis of reasoning traces}
\label{sec:reasoning_traces}
\begin{figure}[h]
\centering
\includegraphics[width=\textwidth]{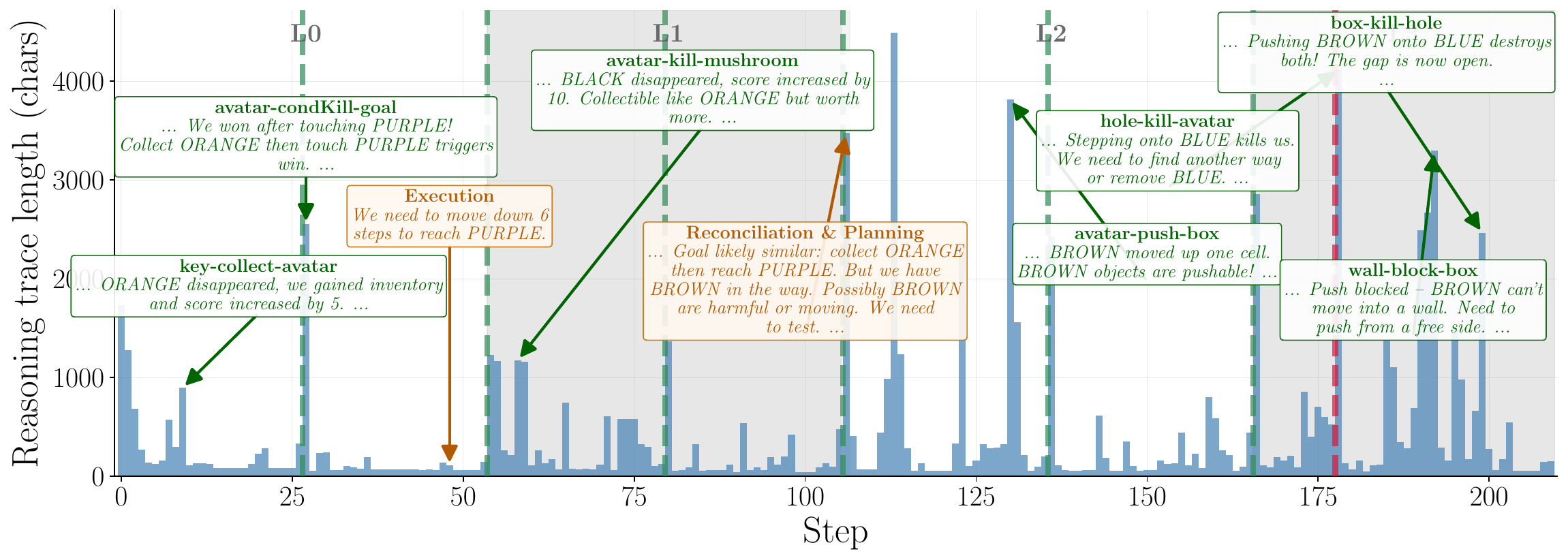}
\caption{\textbf{Example reasoning traces across game steps.}
  DeepSeek~V3.2 on \textsc{Bait} (first four levels, up to first win on L3).
  Bar height shows reasoning trace length (characters) at each step.
  Alternating white/gray shading indicates levels; dashed vertical lines mark episode boundaries (green = win, red = loss).
  Green arrows mark first occurrences of novel game interactions, with curated excerpts from the model's reaction.
  Orange annotations highlight representative execution and planning traces.
  Trace length spikes when an episode starts and at novel interactions, then decays during routine execution.
  Full trace content is explorable in the Interactive Supplementary Material (Appendix~\ref{app:interactive_supp}).}
\label{fig:reasoning_trace}
\end{figure}

The copied-reasoning condition (\S\ref{sec:multiturn}) captures the LRMs hidden chain-of-thought at every game step, providing a direct window into these dynamics. \Cref{fig:reasoning_trace} shows reasoning trace length across the first 210 steps of a single DeepSeek~V3.2 run on \textsc{bait} (4 levels). The trace exhibits a characteristic oscillatory pattern: long traces at the start of each episode and at novel interaction events, decaying to short execution-only traces mid-episode. This pattern repeats at each level and episode boundary and diminishes across levels as the agent accumulates a transferable theory of game mechanics.

Examining the trace content reveals eight distinct reasoning modes, illustrated with verbatim excerpts in Appendix~\ref{app:reasoning_case_study} and explorable step-by-step in the Interactive Supplementary Material (Appendix~\ref{app:interactive_supp}). 

\section{fMRI encoding results}
\label{sec:encoding}
\begin{figure}[h]
    \centering
    \includegraphics[width=\textwidth]{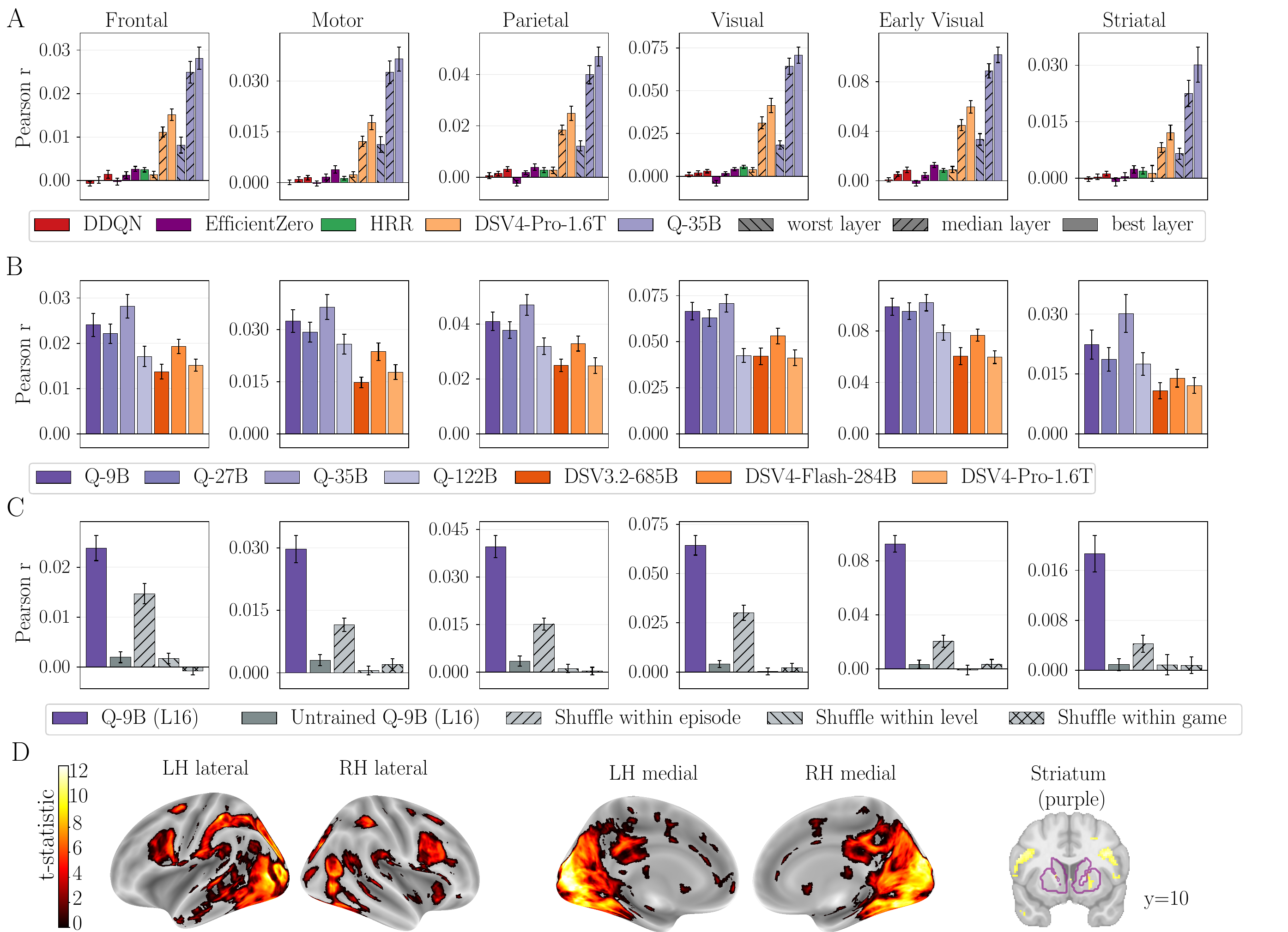}
    \caption{%
    \textbf{Predicting brain activity with frontier LRM embeddings.}
    \textbf{(A)}~Encoding accuracy (per-subject best-layer Pearson $r$,
    averaged across subjects $\pm$ SEM) for RL baselines (DDQN,
    EfficientZero, HRR) compared to the two anchor frontier LRMs
    introduced in \Cref{fig:discovery_combined}: \textbf{DeepSeek V4-Pro}
    (strongest behavioural model) and \textbf{Qwen3.5-35B-A3B} (strongest
    encoding model, identified in panel B), broken out by six functional
    region groups. For multi-layer models, hatched bars show worst- and
    median-layer encoding alongside the best layer.
    \textbf{(B)}~Best-layer encoding accuracy for all seven evaluated
    LRMs (Qwen3.5-397B excluded -- exceeded our self-hosting budget).
    \textbf{(C)}~Comparison to untrained version with matched architecture
    and layer and to three trajectory-shuffle controls that permute the
    temporal organisation of model embeddings within each episode, level,
    or game (computed on Qwen3.5-35B-A3B's best layer).
    \textbf{(D)}~Surface plots for top-performing layer in Qwen3.5-35B-A3B
    using voxelwise one-sample sign-flip permutation test. See
    \Cref{supp:surface_all_llms} for details.%
}
    \label{fig:encoding}
\end{figure}

To compare model representations against human brain activity, we extract hidden-state features while each model passively observes the actual human gameplay trajectories. At each timestep, we feed the LRMs the same observation the human participant saw---formatted identically to the gameplay prompt but without any model-generated reasoning or actions from prior steps---and extract the activation at the final input token, prior to any model output. The context is reset between timesteps, so each extraction is an independent forward pass that captures the model's in-context representation of the current game state. This protocol deliberately isolates representations from reasoning: the LRM encodes the observation but does not plan or act. No rationales appear in context, which gives a controlled feature space comparable across models. The action-only behavioral results (\S\ref{sec:action_only_supp}) validate that models produce meaningful representations under this input format.

We fit encoding models that predict blood-oxygen-level-dependent (BOLD) responses from extracted model features. We use banded ridge regression with four feature bands: (1) main model features, (2) button press indicators, (3) game and level identity, and (4) temporal variables (time within play, time within experiment). Bands 2--4 serve as nuisance regressors, isolating the unique contribution of model representations \citep{nunez2019voxelwise}. We quantify encoding performance as the Pearson correlation between predicted and observed BOLD responses on held-out partitions, averaged across folds. We report results for the anatomically-defined regions of interest (ROIs) used in \citet{tomov2023neural}, grouped into six broad region groups (frontal, motor, parietal, visual, early visual, striatal). Full details are in the Appendix. We repeat this protocol for the DDQN and EfficientZero layers. For EMPA, we reuse the holographic reduced representation (HRR) vectors from \citet{tomov2023neural}, which embed the symbolic EMPA theory in a vector space. Notably, this is far from a complete representation of EMPA, because it only captures EMPA's model-building component but not its planning, exploring, and acting components. We adopt it primarily for direct comparability with prior work on this dataset (see Figure~\ref{supp:hrr_ddqn_repro} for a direct reproduction of their results with our encoding pipeline).


\subsection{LRMs predict brain activity an order of magnitude better than RL baselines}
\label{sec:encoding_headline}

Across all six region groups, frontier LRM representations predict BOLD activity substantially better than the RL and symbolic baselines (Figure~\ref{fig:encoding}A). The gap is largest in visual and early visual cortex, where best-layer encoding accuracy reaches Pearson $r \approx0.07-0.10$ for the strongest LRMs while the best RL baseline (EfficientZero's representation network) sits below $r=0.015$. Smaller but qualitatively similar gaps appear in parietal, frontal, motor, and striatal regions: in every group, LRM representations outperform every RL baseline by roughly an order of magnitude. Although these $r$-values may seem low in magnitude, they fall within the range of those reported in similar prior work on brain encoding during story listening (see \Cref{supp:podcast_bert_reproduction}). The advantage holds for all LRM models we used and replicates using standard ridge regression encoding pipelines with no nuisance regressors (Figure~\ref{supp:pure_ridge}). Trained LRMs consistently outperform random-initialization controls, shuffled-trajectory controls, and all RL baselines across every brain region (\S\ref{supp:controls}, \Cref{supp:cohort_comparison,supp:prompt_ablation,supp:context_ablation,supp:random_init,fig:encoding}). A randomly-initialized LRM produces encoding accuracy comparable to the RL baselines but well below trained LRMs, confirming that the encoding signal reflects learned representations rather than architectural priors (see \Cref{fig:encoding}C or \Cref{supp:random_init} for results across model sizes). Within both the Qwen3.5 and DeepSeek families, encoding accuracy does not scale monotonically with parameter count (\Cref{fig:encoding}B): the Qwen3.5-35B-A3B mixture-of-experts model achieves the highest encoding accuracy in every region group, outperforming both the smaller dense Qwen models and the larger DeepSeek v3.2/v4 models. This decoupling between scale and brain-alignment contrasts with our behavioural results (\Cref{fig:discovery_combined}), which recapitulates work in object recognition where there exists a pointdeeper, more-performant models stop producing performance gains in brain encoding \citep{kubilius2019brain}.

\subsection{Encoding reflects immediate representational structure}
\label{sec:scaffolding}
What is the encoding signal actually measuring? At each timestep, we extract the embedding at the final input token, before any model-generated output, and the context is reset between timesteps (\S\ref{sec:encoding}). We show below this deliberately isolates the model's \emph{in-context representation} of the current game state given recent observations, rather than its planning, reasoning, or accumulated learning.

\textbf{Temporal ordering impacts brain encoding.} Permuting the temporal organization of the embeddings within episode, level, or game substantially reduces encoding accuracy across all region groups (Figure~\ref{fig:encoding}C). This indicates that the signal is locked to the moment-to-moment sequence of human gameplay, not merely to the coarse organizational structure of the experiment. 

\textbf{Prompt-supplied knowledge does not impact brain encoding.} Comparing minimal vs.\ elaborate suggestion levels (\S\ref{sec:multiturn}) yields similar encoding accuracy (Figure~\ref{supp:prompt_ablation}). This suggests that the knowledge in the prompts is not propagating to the LRM's immediate representation of the human game state.

\textbf{Long context does not impact brain encoding.} Truncating the observation history to 50\% or 10\% of its full length preserves encoding accuracy at near-original levels (Figure~\ref{supp:context_ablation}). Brain prediction is not being driven by the LRM's integration of long histories; recent context is sufficient.

Taken together, these results suggest our brain encoding measures \emph{representational alignment} \citep{sucholutsky2023getting}, not alignment of higher-order processes (e.g., learning and planning).

\section{Discussion}
\label{sec:discussion}
We set out to ask whether modern AI systems learn and plan in ways that resemble humans, using a paradigm that affords matched behavioral and brain measurements from the same humans on the same tasks. Two findings define our headline result. Frontier LRMs match human game-solving trajectories where both model-based and model-free deep RL fails. The LRMs also show a discovery--execution gap that humans do not (\S\ref{sec:execution}): once they solve a level, they tend to rigidly re-execute the same trajectory rather than compress to only necessary steps to win the level. Representations from the LRMs also predict human BOLD activity an order of magnitude better than RL baselines across all six functional region groups. Both results are achieved off-the-shelf with a simple system prompt, without fine-tuning, in-context demonstrations, or domain-specific tools. To our knowledge this is the first joint behavioral and neural alignment between humans and LRMs during interactive gameplay.


Two caveats bound the scope of our claims.  First, our interactive paradigm makes estimating per-voxel noise ceilings difficult, since every gameplay trajectory is unique (see Appendix~\ref{app:lessons}). Cross-regional differences in absolute encoding accuracy therefore largely reflect regional signal-to-noise differences. Second, LRMs retain general priors from their opaque training corpora that we cannot control for. We note that every model class in our comparison has access to some form of game-relevant information: deep RL receives direct weight updates from specific levels and EMPA has the ground-truth VGDL ontology and simulator.

A natural puzzle is why a model class trained purely on text predicts brain activity in visual cortex an order of magnitude better than Deep RL agents trained on the gameplay grids themselves. Most of the asymmetry lies in what each model class brings to a new game: LRMs arrive with broad prior knowledge and can acquire each specific game's rules purely in-context. This profile is closer to how humans approach novel games than the from-scratch training of deep RL. Two convergent observations from cognitive science and neuroscience shed light on why text-derived prior knowledge would carry perception-relevant structure: language can be a good proxy for human perceptual judgments \citep{marjieh2022words} and language-evoked representations in the human brain can be distributed widely across cortex, including visual, parietal, and sensorimotor regions \citep{huth2016natural, deniz2019representation, tang2025semantic}.

Language representations capture the structure of \emph{what} the world is, not the cognitive operations that operate over that content \citep{fedorenko2024language}. Although our encoding is capturing representations, we are missing the brain signal associated with active learning and planning itself. Closing this gap will require letting models actively engage with the environment in a way that can align with individual human trajectories, which has two coupled requirements. First, a richer harness, which is a commitment to the model's cognitive architecture: how observations are formatted, what external memory is available, etc. \citep{sumers2023cognitive}. Second, a way to align harnessed gameplay to the human's actual trajectory, since models using an active player harness rather than passive observer can and will diverge from the human trajectories. We expect this combination to recover signal we currently are unable to capture. Frontier LRMs may finally let us test competing accounts of human learning and planning jointly against \emph{both} behavior and the brain.

%

\bibliographystyle{unsrtnat}
\bibliography{neurips_2026}


\section{Acknowledgements}
S.K. is funded by a Leon Levy Fellowship in Neuroscience by the New York Academy of Sciences. R.P.C is funded by a ERC-UKRI Frontier Research Guarantee Starting Grant (EP/Y027841/1). B. Cs. is funded by Department for Science, Innovation and Technology and Pillar VC under the Encode: AI for Science Fellowship.
\clearpage
\appendix

\setcounter{figure}{0} \renewcommand{\thefigure}{S\arabic{figure}}
\setcounter{table}{0} \renewcommand{\thetable}{S\arabic{table}}
\setcounter{equation}{0} \renewcommand{\thetable}{S\arabic{equation}}

\section{Interactive Supplementary Material}
\label{app:interactive_supp}

To accompany the paper we provide an interactive supplementary that lets readers browse every gameplay session reported in this work, scrub through individual replays at any frame, and re-run any game live in the browser with editable rules. The hosted version is available at \url{https://botcs.github.io/reason-to-play/}; the offline bundle works without a server --- open \texttt{catalogue.html} in any modern browser.

\begin{figure}[hbtp]
  \centering
  \includegraphics[width=\linewidth]{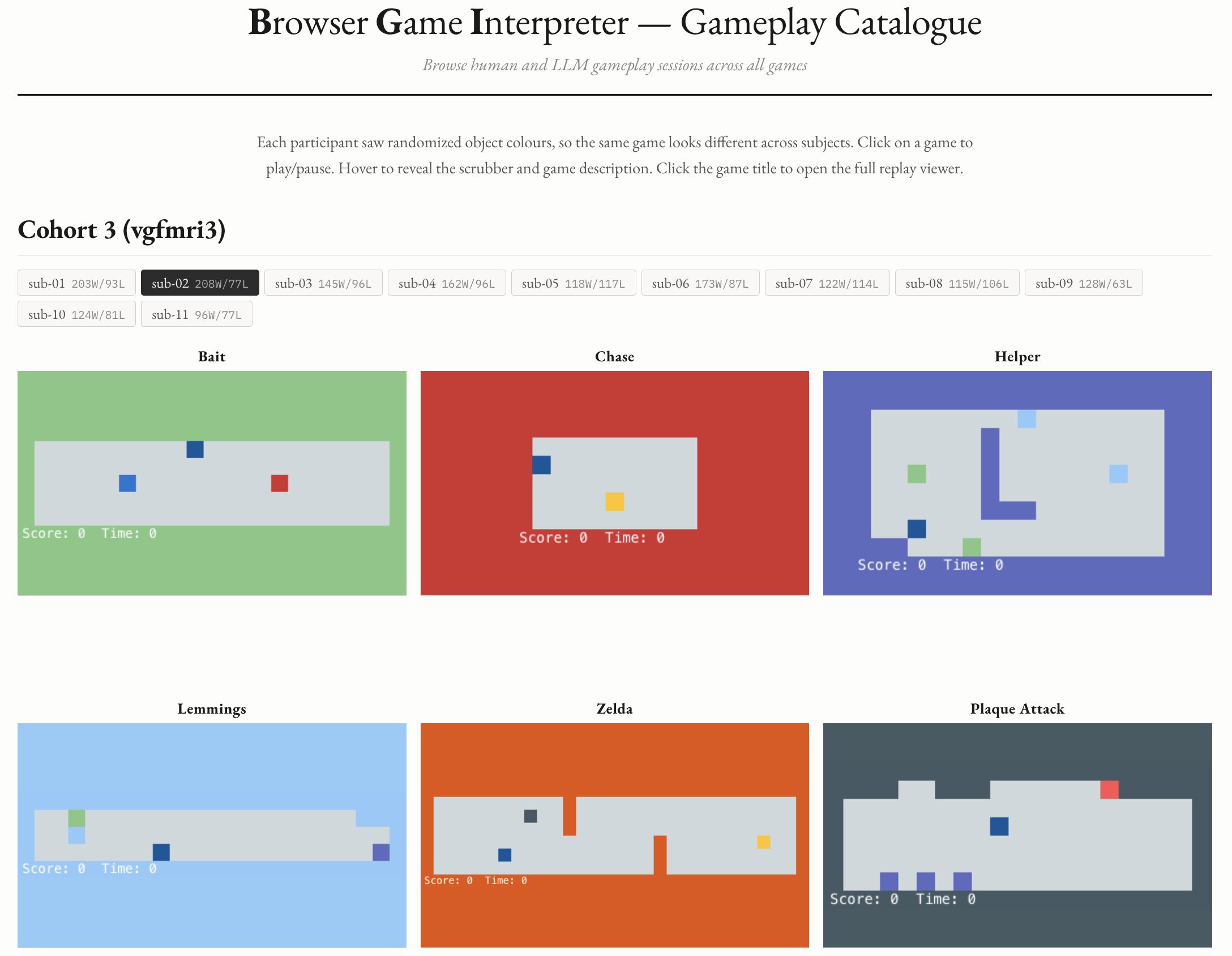}
  \caption{\textbf{Catalogue view.} The active cohort is the second LRM section (action-only mode), with one model selected. Each tile is one game played by that subject; the badges next to each subject tab show win/loss counts. Hovering surfaces the per-tile scrubber and three flap tabs (Game Description, Level Layout, Try Yourself).}
  \label{fig:supp_catalogue}
\end{figure}

\paragraph{Catalogue view (entry point).}
The catalogue (\Cref{fig:supp_catalogue}) is the landing page. It groups gameplay into four cohorts: two human cohorts (\textsc{vgfmri3} with 11 subjects, \textsc{vgfmri4} with 21 subjects) and two LRM cohorts (8 frontier reasoning models, in copied-reasoning and action-only modes). Selecting a subject populates a 6-game grid; clicking a tile autoplays the replay. Hovering reveals a scrubber and three flap tabs that expose the game description, level layout, or an interactive editor.

\begin{figure}[hbtp]
  \centering
  \includegraphics[width=\linewidth]{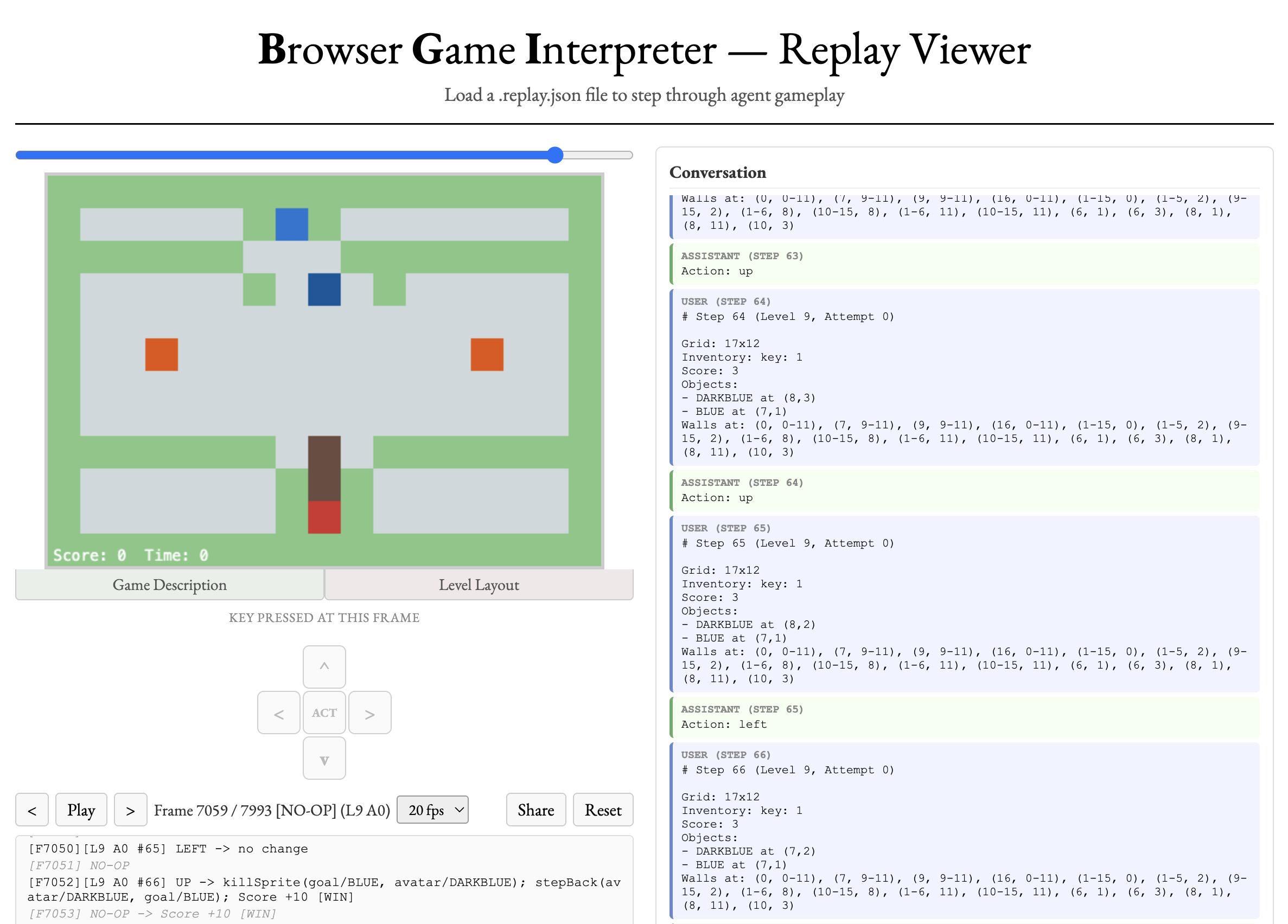}
  \caption{\textbf{Replay viewer.} Left: gameplay canvas with scrubber and step counter. Right: reasoning trace for the current step, action histogram and reasoning-length line plot, and conversation log in raw API format. All panels are step-synchronised.}
  \label{fig:supp_replay}
\end{figure}

\paragraph{Replay viewer (per-step inspection).}
Clicking a game title in the catalogue opens the full replay viewer (\Cref{fig:supp_replay}). It shows synchronised panels for the gameplay canvas, the per-step reasoning trace (full chain-of-thought from copied-reasoning runs), the raw conversation log in OpenRouter API format, and an action histogram with a reasoning-length line plot. Step navigation is via scrubber, keyboard arrows, or click-to-step on the chart. This is the surface we used to study reasoning modes (Appendix~\ref{app:reasoning_case_study}) and trace dynamics (\Cref{fig:reasoning_trace}).

\begin{figure}[hbtp]
  \centering
  \includegraphics[width=\linewidth]{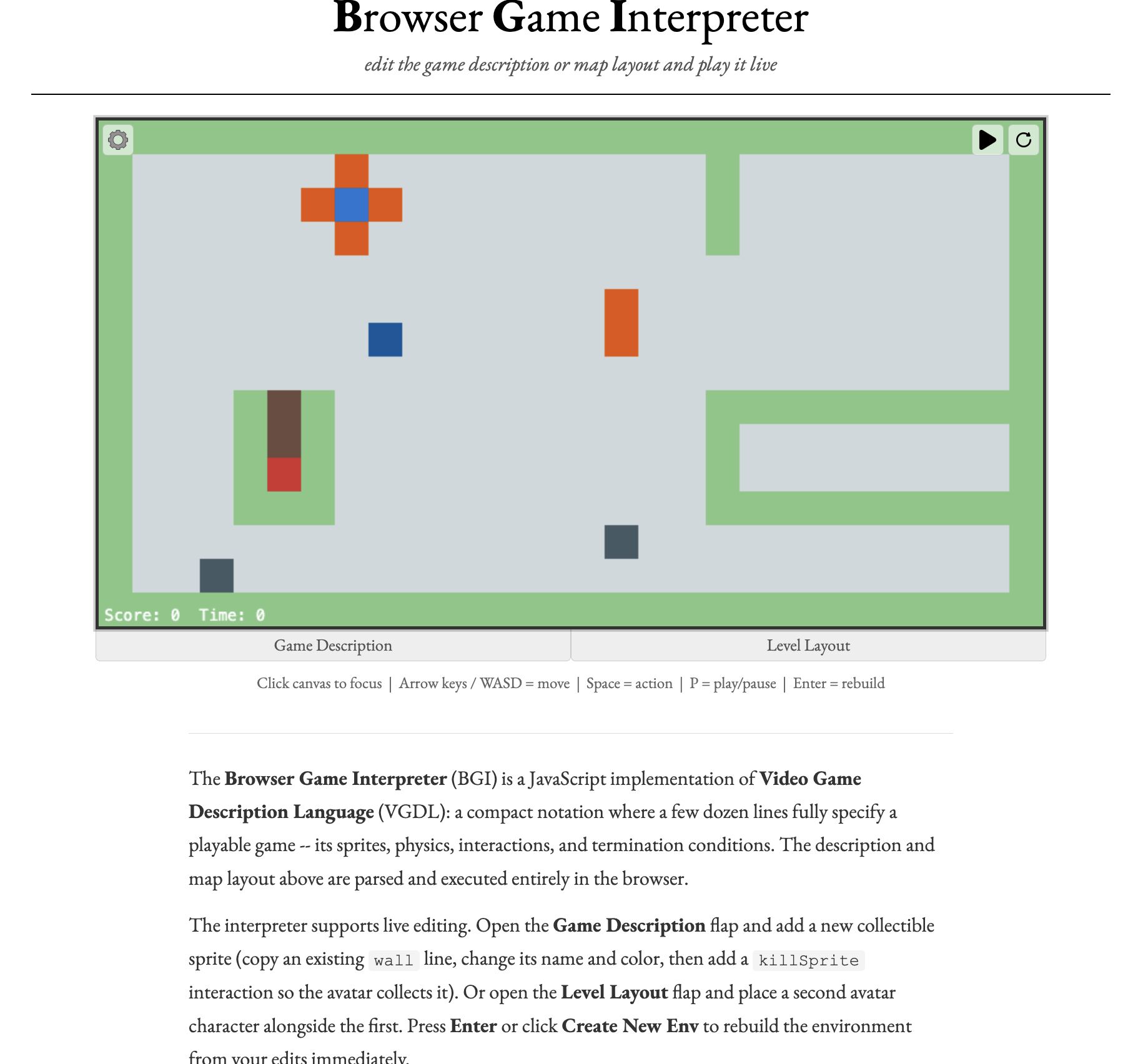}
  \caption{\textbf{Try Yourself view.} The game description and level layout are editable; the canvas re-renders live. Encourages hands-on exploration of how individual VGDL rules shape the resulting environment.}
  \label{fig:supp_interactive}
\end{figure}

\paragraph{Try Yourself (interactive editor).}
The third flap tab on every catalogue card --- ``Try Yourself'' --- opens a live VGDL editor (\Cref{fig:supp_interactive}). The game description and level layout are pre-loaded with the selected subject's exact colour assignments, and both are live-editable. Pressing \texttt{Enter} or clicking ``Create New Env'' rebuilds the environment from the edited text. We encourage reviewers to add a new collectible to the description (copy an existing sprite line, change its colour, add a \texttt{killSprite} interaction) or place a second avatar in the layout, then play with arrow keys / WASD. The same VGDL parser used in our experiments runs directly in the browser, making the semantics of any rule concrete in seconds without reading the parser code.

\section{Supplementary Methods}

\subsection{VGDL-fMRI Dataset Details}
\label{app:dataset}
We use the publicly released VGDL-fMRI dataset of \citet{tomov2023neural}, specifically the \texttt{vgfmri4} cohort, which comprises 21 healthy adult participants (ages 19--36; all right-handed) who learned to play a sequence of grid-world video games while undergoing fMRI in a single $\sim$2-hour session. Each participant completed 6 scanner runs; each run consisted of 3 blocks, with each block containing 3 levels of a single game (9 levels per game total). Each level was played for 60 seconds, restarting immediately on the same level whenever the episode ended in a win or loss. Runs were organized into three partitions with game and level balance preserved across partitions. The games were expressed in the Video Game Description Language \cite{tsividis2021human} and presented in a ``color-only'' rendering mode in which sprites appeared as colored squares with abstract symbols, carrying no semantic priors about the games' rules; participants were instructed that colors and symbols carried no rule information beyond within-game object identity. Participants received a base payment plus a performance-dependent bonus tied to the maximum winning score on a randomly selected level. Across all 32 participants, five games were played in common: \textsc{bait}, \textsc{chase}, \textsc{helper}, \textsc{lemmings}, and \textsc{zelda} (with the sixth game varying between cohorts: \textsc{plaque attack} for participants 1--11 and \textsc{avoid george} for participants 12--32). For all behavioral and neural analyses in this paper, we restrict attention to the four games --- \textsc{bait}, \textsc{chase}, \textsc{helper}, \textsc{lemmings} --- on which all 32 participants played all 9 levels and for which we ran the full set of model agents. The full game-specific rule descriptions, level layouts, and example screenshots are available in the original paper \citep{tomov2023neural} and at the project repository \url{https://github.com/tomov/RC_RL/tree/fmri/fmri_all_games}.

\subsection{Game Descriptions}
\label{app:games}
\begin{figure}[h]
    \centering
    \includegraphics[width=\textwidth]{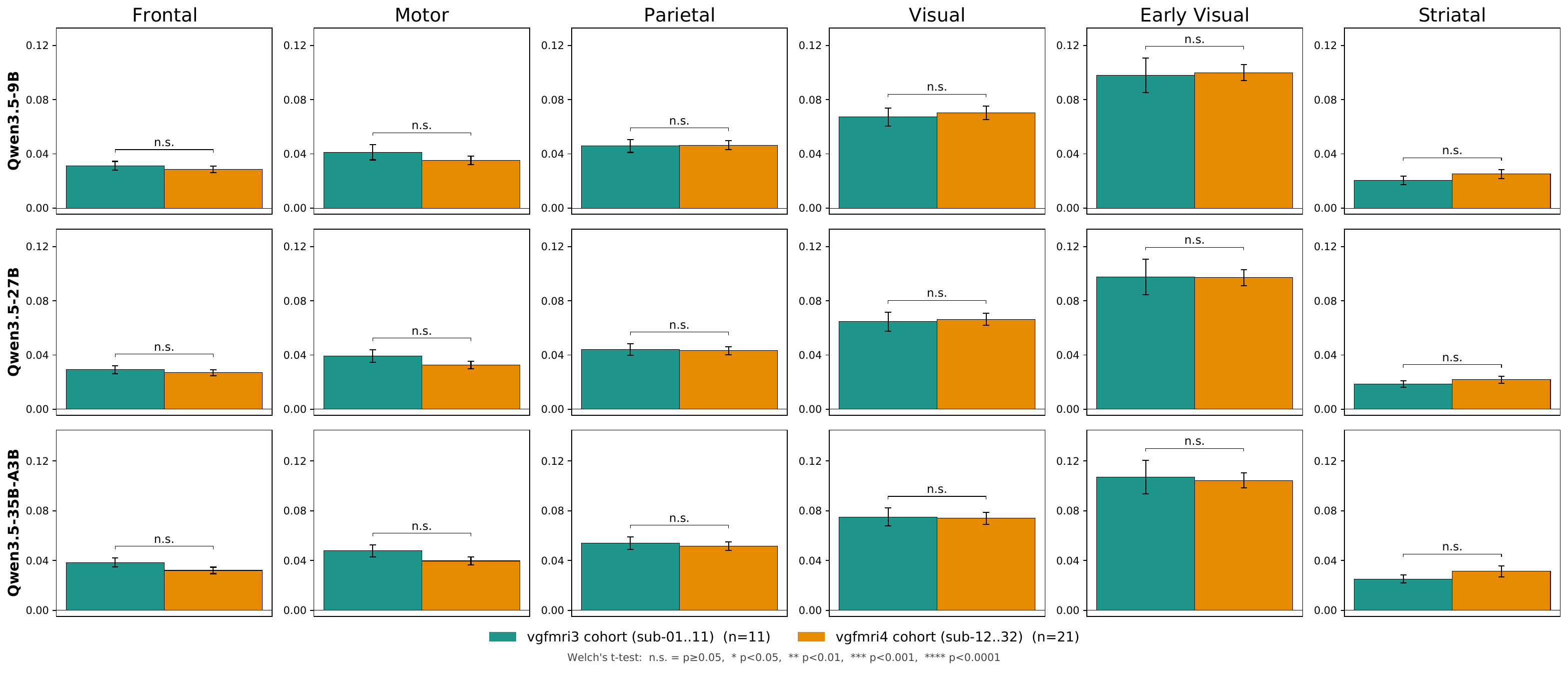}
    \caption{%
        \textbf{Mean encoding accuracy within region groups is statistically indistinguishable
        between the vgfmri3 and vgfmri4 cohorts.}
        Per-subject best-layer Pearson $r$ ($\pm$ SEM) for the three
        Qwen3.5 models that were run on both cohorts (rows: 9B, 27B,
        35B-A3B; columns: six anatomical region groups), comparing
        the vgfmri3 cohort (sub-01--sub-11, $n=11$, teal) to the
        vgfmri4 cohort (sub-12--sub-32, $n=21$, orange). Welch's
        independent two-sample $t$-tests between cohorts find no
        significant difference (n.s.\ at $p<0.05$) in any of the 18
        (model, region) panels. This is to be expected due to the high overlap in the cohorts' set of games ($5$ games out of $6$ were the same).%
    }
    \label{supp:cohort_comparison}
\end{figure}

The games we analyze are 2D grid-world games written in the Video Game Description Language \citep{tsividis2021human}. They are taken from the game set used in \citet{tomov2023neural} and were originally selected by \citet{tsividis2021human} to span a heterogeneous space of mechanics --- including pushing and obstruction, agent-tracking pursuit, multi-step subgoaling, hazard avoidance, ranged interception, and cooperative guidance --- so that a learning agent must infer qualitatively different rule structures across games. All games share the same 5-action interface (left, right, up, down, action) and the same color-anonymized presentation described in Appendix~\ref{app:dataset}; the only information distinguishing one sprite from another within a game is its color and its grid position. The full VGDL game specifications and per-level layouts are available at \url{https://github.com/tomov/RC_RL/tree/fmri/fmri_all_games}; here we provide brief informal summaries to support interpretation of the results.\paragraph{Game set and the \textsc{vgfmri3}/\textsc{vgfmri4} variants.}
Following \citet{tomov2023neural}, each participant played six games over the course of the scanner session, each consisting of nine levels of increasing complexity. Five of the six games are identical across all participants: \textsc{bait}, \textsc{chase}, \textsc{helper}, \textsc{lemmings}, and \textsc{zelda}. The sixth game differs between two dataset variants: participants in \textsc{vgfmri3} (sub-1 through sub-11) played \textsc{plaque attack}, while participants in \textsc{vgfmri4} (sub-12 through sub-32) played \textsc{avoid george} in its place. Behavioral analyses (\S\ref{sec:behavior}) include all participants from both variants (n=32n=32
n=32); brain-encoding analyses are restricted to participants from the vgfmri4 cohort ($n=21$) as we found results across the two cohorts to be statistically indistinguishable (see Figure~\ref{supp:cohort_comparison}). For each game, later levels introduce sprite types or interaction outcomes not present in earlier levels, ensuring that the agent (human or model) continues to encounter novel rule-discovery opportunities throughout the 9-level sequence rather than reaching a steady state after the first level or two. This design choice is crucial for the brain encoding analyses, since it ensures the human BOLD signal contains learning-related variance throughout each scanner run.\paragraph{\textsc{bait}.} A Sokoban-like puzzle game. The avatar moves on a grid containing walls, holes that kill the avatar on contact, pushable boxes, a key, and a goal door. Holes can be neutralized by pushing a box into them, after which both the box and the hole disappear and the avatar can safely traverse the cell. The level is won when the avatar collects the key and reaches the door, and lost if the avatar contacts an unfilled hole. Later levels introduce dirt cannons that fill any holes in the cannon's path (so filling the wrong holes can render the level unwinnable), and replace the key-and-door mechanic with a manufacturing step in which the avatar must push metal into a mold to forge the key. These additions require multi-step planning over which holes to fill in which order and over which sub-goals to pursue first.\paragraph{\textsc{chase}.} A pursuit game. The avatar must catch all the birds on the grid; the birds flee from the avatar's current position on each step. Touching a bird turns it into a carcass. If a fleeing bird later touches a carcass, the carcass transforms into a predator bird that pursues and kills the avatar --- introducing a sequential subgoaling structure in which the order of catches matters. Later levels introduce gates that sporadically release predator birds, alternative outcomes in which touching a bird kills it directly, and additional NPC types (sheep that pace on fixed paths, zombies that chase the avatar after triggering events). The level is won when all targetable sprites have been caught.\paragraph{\textsc{helper}.} A cooperative game. The avatar must help one or more NPC ``minions'' reach their food. The food may be guarded by a boiling pot of water that only the avatar can destroy, or the minions may be boxed in by a fence that the avatar must push food through to feed them. Later levels invert the win condition (the avatar wins by eating all the food, but only after delivering it to the minions for processing) and introduce ranged actions (shooting a path through colored fences). The level is won when all minions have been fed; failure modes include the avatar contacting hazardous sprites. The cooperative structure means the avatar's actions need to be coordinated with the predicted trajectories of the minions rather than only with the avatar's own goal.\paragraph{\textsc{lemmings}.} A guidance game inspired by the 1991 puzzle game of the same name. NPC lemmings emerge from a spawn point and walk autonomously through the level, typically toward hazards (e.g.,\ pits) that kill them on contact. The avatar must intervene by shoveling tunnels through dirt to redirect the lemmings toward a safe destination. Each shovel action incurs a small score penalty. Later levels introduce a mole NPC that the avatar can release to clear dirt automatically (with the side-effect that the mole kills the avatar on contact, and in some levels transforms into a snake that chases the avatar if it touches a lemming). The level is won when a sufficient number of lemmings reach the destination and lost if too many die.\paragraph{\textsc{zelda}.} An exploration-and-combat game inspired by the original \emph{Legend of Zelda}. The avatar navigates a grid containing walls, randomly-moving enemy creatures (which kill the avatar on contact), a key, and a goal door. The avatar carries a sword that can be used (via the action key) to destroy enemy creatures. The level is won when the avatar collects the key and reaches the door. Later levels increase the number of required keys (up to three), introduce mobile keys and doors that move randomly, and introduce levels in which keys and doors are carried by NPC elves that flee the avatar --- transforming the task from spatial planning into a pursuit problem nested inside the basic key-and-door structure.\paragraph{\textsc{avoid george} (\textsc{vgfmri4}).} A defense game in which the avatar must keep a population of citizen NPCs calm. Citizens become annoyed over time; if every citizen is annoyed and there are no calm citizens left, the avatar loses. The avatar can feed candy to annoyed citizens to calm them down. A hostile NPC, George, roams the grid; if George touches the avatar, the avatar dies. The level is won by keeping at least one citizen calm for 500 game steps. Later levels modify the candy mechanic so that feeding candy makes annoyed citizens disappear (with the win condition becoming ``all annoyed citizens disappear''), allow the avatar to throw candy at range, introduce levels in which the avatar starts trapped in a sealed room and must tunnel out before acting, and progressively increase George's movement speed.\paragraph{\textsc{plaque attack} (\textsc{vgfmri3}).} A defense game inspired by the Atari 2600 game of the same name. Burger and hotdog enemies emerge from spawn points and advance toward a row of vulnerable cavity sprites; if the enemies reach the cavities, the avatar dies. The avatar must intercept the advancing enemies by shooting them with a laser. The base level requires destroying all 40 burgers and hotdogs to win. Later levels reduce the enemy count but speed them up while adding a roving drill that must be avoided, replace the drill with bouncing projectiles that kill the avatar on contact, and introduce an alternative win condition in which the avatar destroys gold filling sprites instead of the enemies.
\subsection{DDQN Training Protocol}
\label{app:ddqn}

We use a Double DQN \citep{van2016deep} as a model-free deep RL baseline. Our pipeline forks the implementation released with \citet{tomov2023neural} (\url{https://github.com/tomov/RC_RL}), with seven implementation bugs corrected (Appendix~\ref{app:ddqn_bugs}) and two methodological changes detailed below: per-game hyperparameter tuning, and curriculum-based training over the full 9-level sequence. In contrast to the single shared configuration used in \citet{tomov2023neural}, we conducted an extensive per-game hyperparameter search; we view this as raising the performance bar for the model-free baseline so that any remaining gap to the LRMs reflects a genuine capability difference rather than an undertrained baseline.

\paragraph{Architecture and inputs.}
The Q-network is the same 3-convolutional-layer + 2-fully-connected-layer architecture used in \citet{tomov2023neural}, with the input representation replaced as described in Appendix~\ref{app:state_repr_fixes}: each sprite type is assigned a dedicated binary channel, yielding an observation tensor of shape $(C \times H \times W)$ where $C$ is the number of distinct sprite types in the game and $H, W$ are the native grid dimensions of each level. Frames are stacked along the channel axis (stack size selected per game by the sweep). The action space is the same 5-action interface used by humans plus a no-op, totalling 6 actions.

\paragraph{Curriculum-based training.}
In contrast to the original sequential overfitting regime---250k on-policy gradient updates per level, looping through the 9 levels for 100 epochs---we train a single agent across the full 9-level curriculum with a fixed budget of 100k gradient updates per level (900k updates per training run). Levels are presented in ascending order; the agent advances when it achieves a per-level mastery threshold or exhausts the per-level update budget, whichever comes first. Replay buffer, target network, and exploration schedule are persisted across levels so that the agent retains learned representations as the curriculum progresses, allowing us to capture before/after-learning representations at each level boundary for the encoding analyses.

\paragraph{Per-game hyperparameter sweep.}
For each of the 12 (cohort, game) pairs we ran an independent Bayesian-optimisation sweep using \texttt{wandb} sweeps with Hyperband early-stopping. The search budget was 256 configurations evaluated across 4 successive-halving stages per game (configurations that underperform on the early-stage objective are pruned and the surviving configurations continue to the next training-budget tier). The search ranges over the following hyperparameters:
\begin{itemize}
\item Learning rate: $\{10^{-5},\, 2.5{\cdot}10^{-5},\, 5{\cdot}10^{-5},\, 10^{-4},\, 2.5{\cdot}10^{-4},\, 5{\cdot}10^{-4}\}$
\item Discount $\gamma$: $\{0.97,\, 0.985,\, 0.995,\, 0.999\}$
\item Replay buffer size: $\{10^{5},\, 3{\cdot}10^{5},\, 10^{6}\}$
\item Mini-batch size: $\{32,\, 64,\, 128\}$
\item Target-network update frequency: $\{1{,}000,\, 5{,}000,\, 10{,}000,\, 30{,}000\}$ steps
\item $\epsilon$-decay schedule length: $\{10^{5},\, 3{\cdot}10^{5},\, 10^{6}\}$ steps
\item Final $\epsilon$: $\{0.01,\, 0.02\}$
\item Gradient-norm clip: $\{0,\, 5,\, 10\}$
\item Frame-stack size: $\{4,\, 8\}$
\end{itemize}
The sweep objective is the level-averaged episodic reward across the 9-level curriculum. The configuration that maximises this objective is retained per game and used for all downstream behavioural and encoding analyses. Sweep YAML files and the resulting per-game configurations are released alongside the supplementary code.

\subsection{EfficientZero Training Protocol}
\label{app:ez}

We additionally evaluate EfficientZeroV2 (EZv2; \citealt{wang2024efficientzero}), a model-based deep RL agent that learns a latent dynamics model and plans via Monte Carlo tree search (MCTS). Training and feature extraction follow the EZv2 reference implementation released by the original authors, with two modifications mirroring the changes we made to the DDQN protocol: a per-game warmup curriculum, and an extensive per-game hyperparameter sweep that replaces the original work's single shared configuration.

\paragraph{Architecture and inputs.}
EZv2 comprises four learned modules: a representation network that encodes the observation into a latent state, a dynamics network that predicts the next latent state and reward conditional on an action, a value head, and a policy head. The representation network ingests the same per-sprite-type channel observation used by the DDQN baseline (Appendix~\ref{app:state_repr_fixes}), so that any difference between DDQN and EZv2 reflects the model-free vs.\ model-based learning regime rather than perceptual preprocessing. For the encoding analyses we extract hidden states from all internal layers of the four modules.

\paragraph{Warmup curriculum.}
For each game we extend the standard 9-level VGFMRI curriculum with three additional game-specific warmup levels prepended to level~0, yielding a 12-level training schedule. The warmup levels strip out late-game sprite types and reduce grid complexity while preserving each game's core interactions; their role is to bootstrap EZv2's world model on simpler dynamics before exposing it to the full evaluation levels. A single set of warmup levels is hand-designed per game and re-used across all hyperparameter configurations. The agent is trained sequentially through the 12-level extended curriculum, advancing on mastery or budget exhaustion analogously to the DDQN protocol. For all behavioural and encoding analyses we report metrics only on the 9 evaluation levels (0--8); steps and gradient updates spent on the warmup levels are accounted for separately and excluded from the cumulative-step axis used in human-comparable behavioural plots, except where noted via the \texttt{--include-curriculum} flag (Appendix~\ref{app:pergame_behavior}).

\paragraph{Per-game hyperparameter sweep.}
As with the DDQN baseline, we conducted an independent per-game Hyperband sweep rather than reusing a single shared configuration across games. For each of the 12 (cohort, game) pairs we evaluated 256 configurations across 4 successive-halving stages per game, ranging over the EZv2 learning rate, discount $\gamma$, replay-buffer size, mini-batch size, target-network update frequency, exploration-temperature schedule, MCTS simulation count, frame-stack size, and random seed. The configuration that maximises level-averaged episodic reward across the 9 evaluation levels is retained per game and used for all downstream behavioural and encoding analyses.



\subsection{fMRI Preprocessing}
\label{app:preprocessing}

We applied an SPM12-style preprocessing pipeline to the BIDS-formatted data, modeling after the procedure of \citet{tomov2023neural}. Initial preprocessing was performed with fMRIPrep \citep{esteban2019fmriprep}, which handles motion correction, susceptibility distortion correction, coregistration to the anatomical reference, and normalization to the MNI152NLin2009cAsym template at 2~mm isotropic resolution. We then applied a downstream pipeline implemented in Python using \texttt{nilearn} to produce voxelwise timeseries ready for encoding analysis. The pipeline applies the following steps in order:

\paragraph{Spatial smoothing.} Each preprocessed BOLD run is smoothed with an 8~mm FWHM Gaussian kernel, increasing signal-to-noise for distributed cognitive representations while preserving region-level spatial precision.

\paragraph{Confound regression.} We regress out 14 nuisance regressors per run: the 6 rigid-body motion parameters (3 translation, 3 rotation), their first temporal derivatives, and the mean signal in CSF and white-matter masks (computed by fMRIPrep). Confound regression is performed jointly with high-pass filtering and detrending in a single weighted-least-squares step using \texttt{nilearn.image.clean\_img}, which avoids the bias that arises from sequential application of these operations.

\paragraph{Temporal filtering.} We apply a high-pass filter with a 1/128~Hz cutoff (period $\approx$~128~s) to remove slow drifts arising from scanner instabilities, and a linear detrend to remove run-level linear trends.

\paragraph{AR(1) pre-whitening.} To correct for temporal autocorrelation in the BOLD signal, we apply SPM12-style autoregressive pre-whitening. We estimate a global AR(1) coefficient $\rho$ as the median across voxels of the lag-1 Pearson autocorrelation of the cleaned timeseries, then apply the whitening transform $y'[t] = y[t] - \rho \cdot y[t-1]$ to each voxel. The first timepoint of each run is dropped as a result. This step renders the residual noise approximately temporally independent, satisfying the i.i.d.\ assumption of the encoding model's regression objective.

\paragraph{Voxel-wise z-scoring.} Each voxel's timeseries is z-scored within run, so that the encoding model's regression coefficients reflect partial correlations rather than being driven by mean-signal differences across voxels or runs.

The resulting per-(subject, run) voxelwise data are saved alongside the brain mask and the affine transform required to reconstruct results in MNI space. All preprocessing parameters --- including the AR(1) coefficient estimated for each run --- are retained as metadata for reproducibility. Code for the preprocessing pipeline is available in the supplementary materials.

\subsection{Encoding Model}
\label{app:encoding}

We fit voxelwise encoding models that predict each subject's preprocessed BOLD timeseries (Appendix~\ref{app:preprocessing}) from computational model features extracted on the human gameplay trajectories that subject observed. The pipeline operates in two stages: (i) per-trajectory alignment of model features to BOLD volumes, and (ii) banded-ridge regression with cross-validation. Reported encoding accuracies in the main text are the held-out Pearson correlation between predicted and observed BOLD, averaged across cross-validation folds.

\paragraph{Feature extraction and alignment.}
For each model and each (subject, run, episode) tuple, model features are extracted at every game-state observation. Game-state observations occur asynchronously with respect to the scanner TR, so we resample features onto the BOLD grid using the behavioral timestamps recorded during scanning. For each TR, features from all observations falling within that TR are averaged, yielding a $(n_{volumes} \times n_{features})$ design matrix. For LRMs, where features may be subsampled relative to the gameplay frame rate, we align using each LRM observation's own timestamp; states whose timestamps fall outside the AR(1)-whitened time window are dropped. Alignment is verified by checking that model timestamps match the behavioral timestamps in the episode record to within $100$~ms. We additionally extract three nuisance feature groups for each episode: a 5-channel keypress indicator (one channel per action key), a one-hot game and level identity, and two scalar time regressors (time within the current episode, time within the experiment).

\paragraph{Cross-validation partitioning.}
We use leave-one-partition-out cross-validation with three partitions corresponding to the three within-game level groups (levels 0--2, 3--5, 6--8). Each partition contains levels from every game, so train and test sets are matched on game identity and differ only in which levels are held out. Each fold trains on two partitions and predicts the held-out one. This setup also ensures temporal separation of the training and test sets in each fold (see \Cref{fig:overview}B). Encoding accuracy is the Pearson $r$ between predicted and observed BOLD on the held-out partition, averaged across the three folds. All preprocessing steps (z-scoring, PCA when applicable, ridge fitting) are computed on training data only and applied to test data.

\paragraph{Lagged design matrix.}
To accommodate the hemodynamic delay,  we lag each feature group by $\{2, 3, 4, 5\}$ TRs into the past, so that $\mathrm{BOLD}[t]$ is predicted by features from $t-5$ through $t-2$ (the "Finite Impulse Response" (FIR) approach, see \citet{kay2008modeling,huth2016natural,de2017hierarchical}). Lags are applied within episode boundaries (no information leaks across plays). When the lag-expanded feature count would exceed the training-set size (which is the case for all our models except for a few low-dimensional layers in EfficientZero) we apply PCA to the unlagged features first, retaining the same number of components across models equal to $N_{TRs}/N_{lags}$ ($N_{lags}=4)$, effectively matching the dimensionality of the regressors across all of our models. PCA is fit on the training fold only; the test fold is projected using the train-fold loadings.

\paragraph{Banded ridge regression.}
We fit the encoding model using banded ridge regression \citep{nunez2019voxelwise}, which assigns a separate regularization penalty to each feature band. This allows the model to weight each band according to its predictive value rather than forcing a single regularization strength across heterogeneous predictors. Our four feature bands are: (1) main model features (the model layer being evaluated), (2) game and level identity, and optionally (3) keypress indicators and (4) time regressors. Bands 2--4 act as nuisance regressors, soaking up variance attributable to motor activity, run/level structure, and slow temporal trends so that the main band's predictive contribution reflects representational structure rather than these confounds. We use the implementation in the \texttt{himalaya} library \citep{dupre2025voxelwise} with the random-search solver, $20$ alpha values logarithmically spaced over $[10^{-5}, 10^{10}]$, and $100$ random-search iterations per fold. Hyperparameter selection within each training fold uses a nested leave-one-partition-out CV.

\begin{figure}[h]
    \centering
    \includegraphics[width=\textwidth]{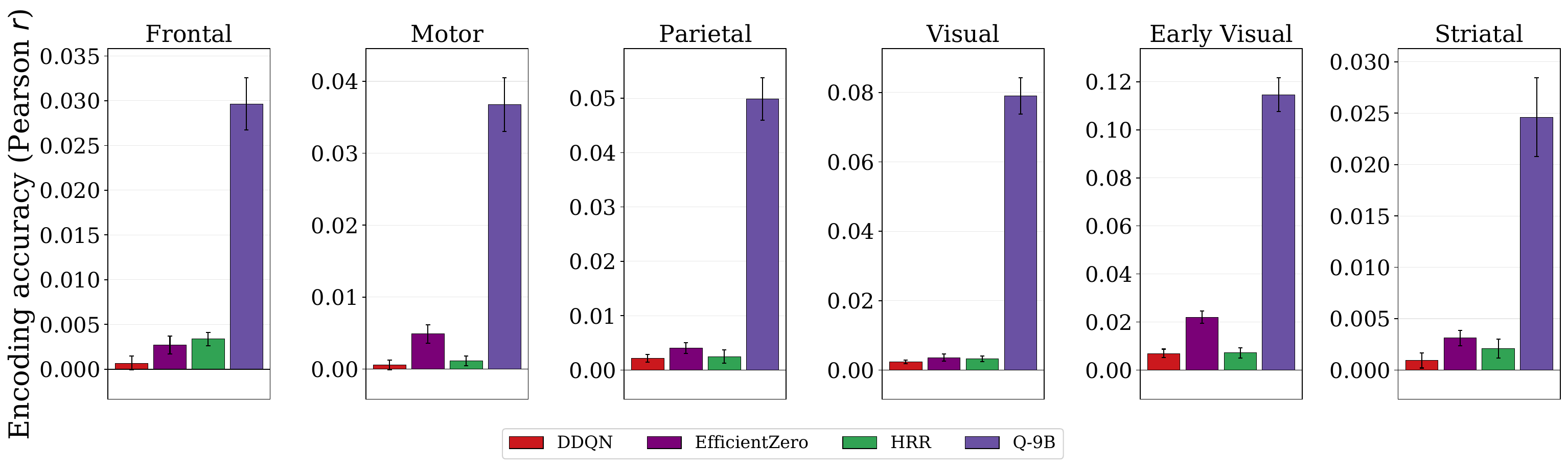}
    \caption{%
        \textbf{The LRM--baseline encoding gap is preserved under
        plain-ridge encoding without nuisance regression.}
        Encoding accuracy (per-subject Pearson $r$, $\pm$ SEM across
        $n=21$ subjects from the vgfmri4 cohort) within six anatomical
        region groups for the three RL-style baselines (DDQN,
        EfficientZero, HRR) and a representative LRM (Qwen3.5-9B at
        layer 16). All models are fit with plain ridge regression on a
        single feature band with no
        banded ridge and no motion, level-identity, or time nuisance
        regressors. The headline result from the main text holds with this more minimal encoding pipeline:
        Qwen3.5-9B's encoding accuracy exceeds every RL baseline by roughly
        an order of magnitude in every region group. %
    }
    \label{supp:pure_ridge}
\end{figure}

\paragraph{Reporting marginal performance.}
Banded ridge regression produces a prediction of the target that can be decomposed into individual contributions by each feature band. For each LRM and baseline RL model, we report the predictive performance attributable specifically to the main band by isolating the contribution of the main band and evaluating it against the held-out BOLD. This yields, per voxel and per fold, a marginal performance per band. The values plotted in our main results are the main-band (LRM, Deep RL, or HRR model features) marginal performance averaged across folds. Reporting the marginal is critical for the cross-model comparisons in this paper: it isolates the unique variance explained by the model's representations from variance explained by motor activity, game/level identity, and time, rather than crediting the model for variance any reasonable predictor would absorb. Note that the original work \citet{tomov2023neural} took a similar approach by controlling for game identity as a nuisance regressor and we take a more rigorous approach here. We also replicated our main effects in a minimal ridge regression paradigm where there are no nuisance regressors (\Cref{supp:pure_ridge}).

\section{Supplementary Results --- Behavior}

\subsection{Per-Game Behavioral Distributions}
\label{app:pergame_behavior}

\begin{figure}[h]
\centering
\includegraphics[width=\textwidth]{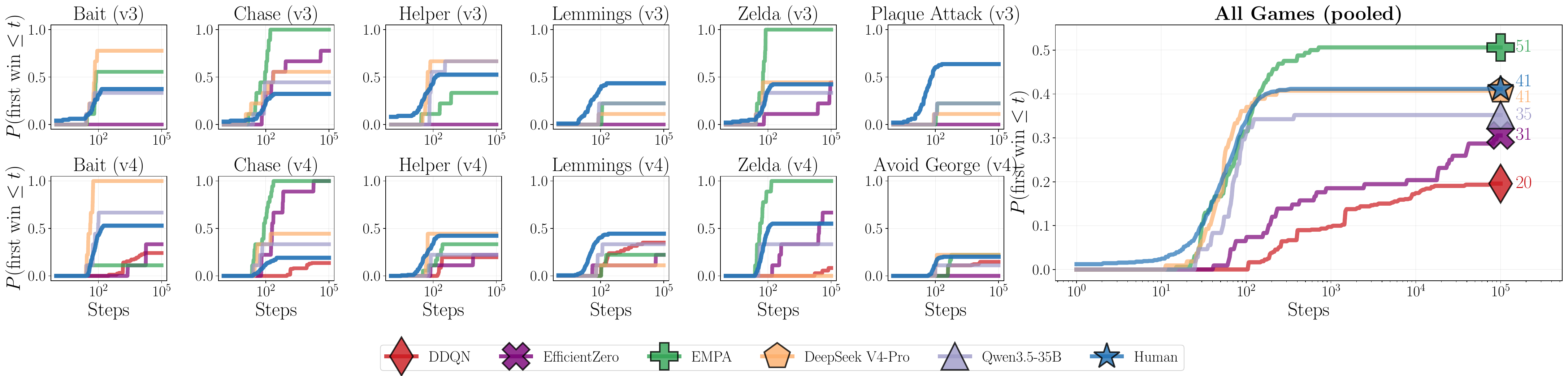}
\caption{\textbf{Kaplan--Meier survival curves (baselines).}
  Per-game panels and pooled aggregate for Human, DDQN, EfficientZero,
  and the two anchor LRMs (DeepSeek V4-Pro, Qwen3.5-35B-A3B).
  The $y$-axis shows $P(\mathrm{level\ solved\ by\ step\ }t)$.}
\label{fig:km_cross_paradigm_supp}
\end{figure}

\begin{figure}[h]
\centering
\includegraphics[width=\textwidth]{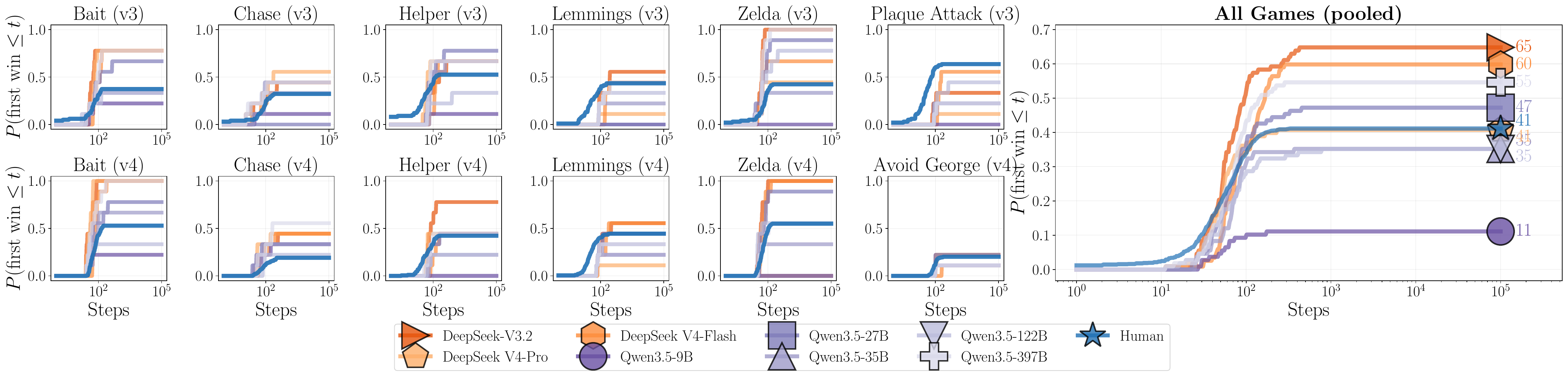}
\caption{\textbf{Kaplan--Meier survival curves (all LRMs).}
  Same format as \Cref{fig:km_cross_paradigm_supp} for all eight frontier LRMs.}
\label{fig:km_llms_supp}
\end{figure}

\begin{figure}[h]
\centering
\includegraphics[width=\textwidth]{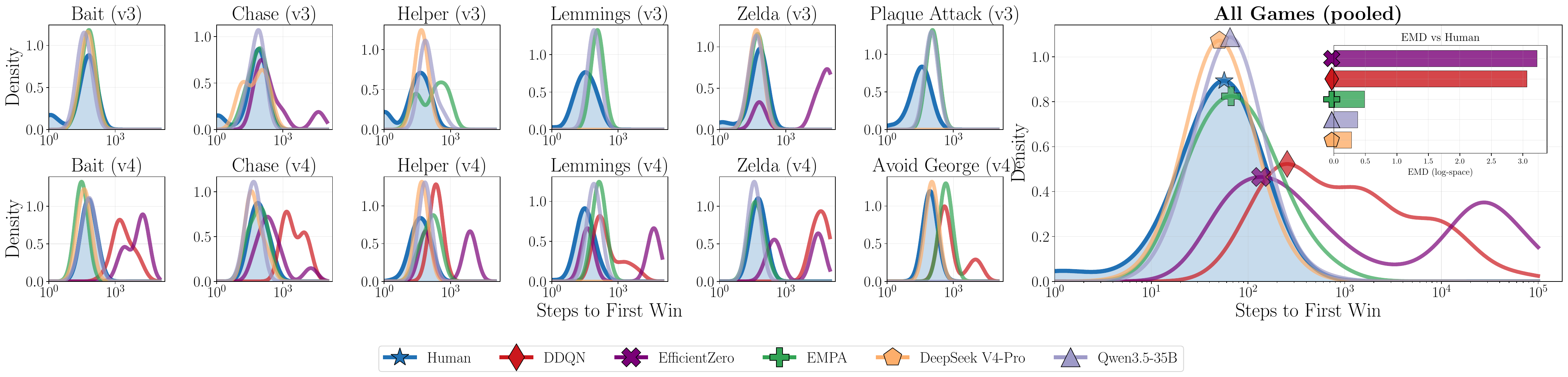}
\caption{\textbf{Per-game discovery time distributions (baselines).}
  Peak-normalized KDE of cumulative steps to first win per level (log scale).
  Human distribution shown as shaded fill; agents as curves.}
\label{fig:discovery_cross_paradigm_supp}
\end{figure}

\begin{figure}[h]
\centering
\includegraphics[width=\textwidth]{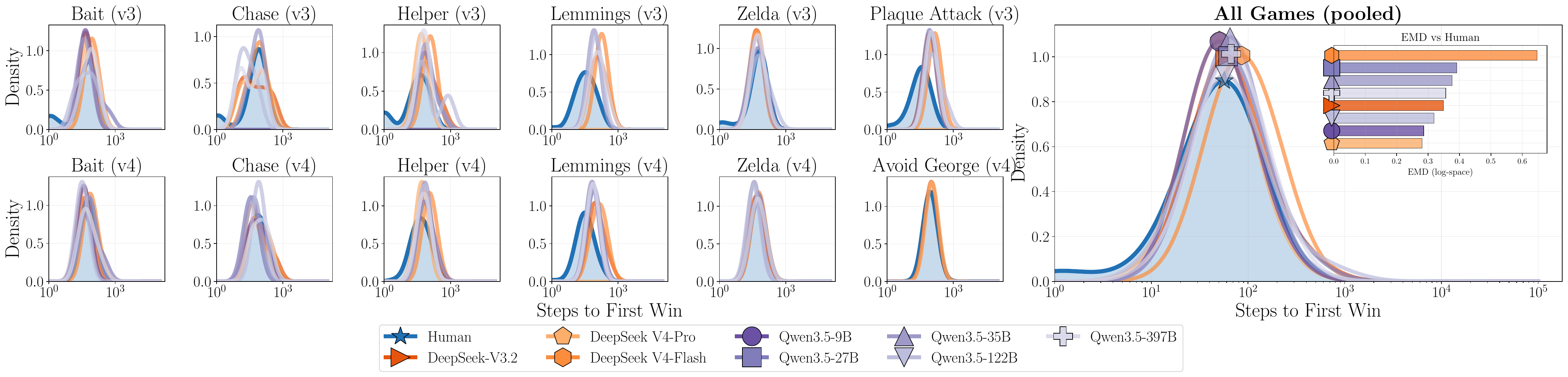}
\caption{\textbf{Per-game discovery time distributions (all LRMs).}
  Same format as \Cref{fig:discovery_cross_paradigm_supp} for all eight frontier LRMs.}
\label{fig:discovery_llms_supp}
\end{figure}

\begin{figure}[h]
\centering
\includegraphics[width=\textwidth]{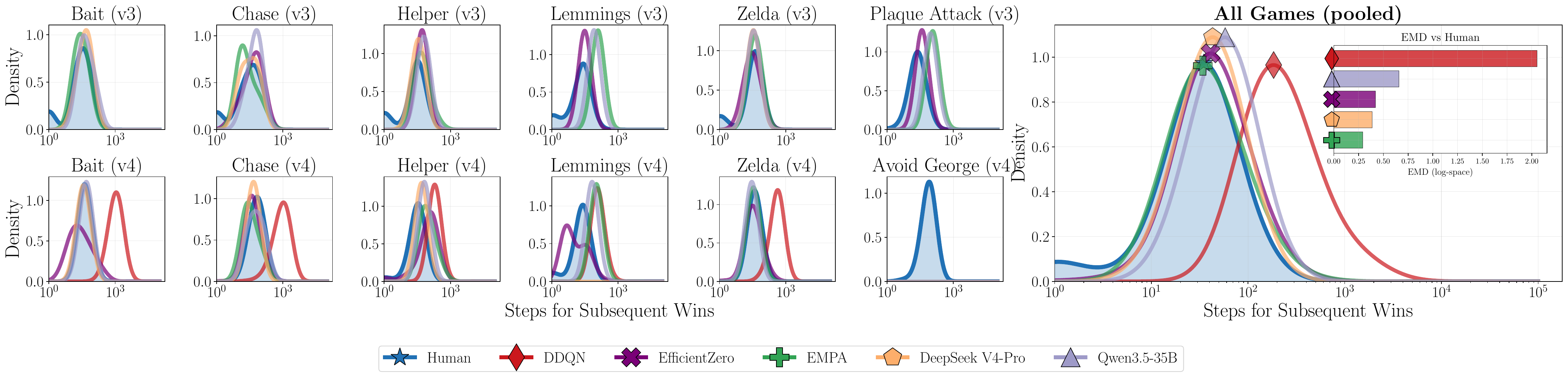}
\caption{\textbf{Per-game execution time distributions (baselines).}
  Peak-normalized KDE of steps per win after the first on the same level.
  The spike near $10^0$ in the human distribution reflects single-keypress
  wins on trivially easy levels.}
\label{fig:execution_cross_paradigm_supp}
\end{figure}

\begin{figure}[h]
\centering
\includegraphics[width=\textwidth]{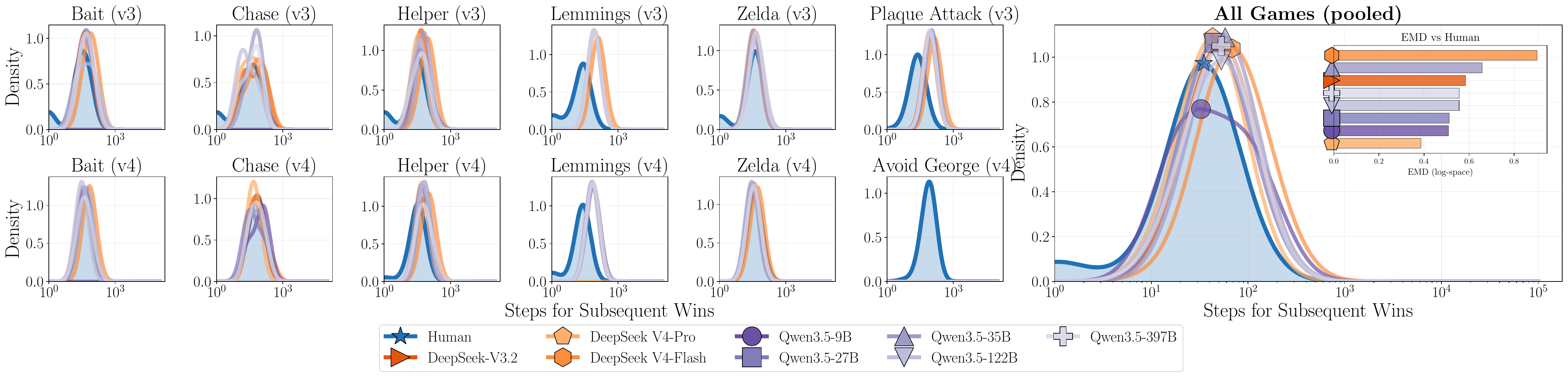}
\caption{\textbf{Per-game execution time distributions (all LRMs).}
  Same format as \Cref{fig:execution_cross_paradigm_supp} for all eight frontier LRMs.}
\label{fig:execution_llms_supp}
\end{figure}

\Cref{fig:km_cross_paradigm_supp,fig:km_llms_supp} show Kaplan--Meier survival curves (fraction of levels solved as a function of cumulative steps, \citep{kaplan1958nonparametric}) broken down by game and cohort. \Cref{fig:discovery_cross_paradigm_supp,fig:discovery_llms_supp} show per-game discovery time distributions (KDE of cumulative steps to first win). \Cref{fig:execution_cross_paradigm_supp,fig:execution_llms_supp} show per-game execution time distributions (KDE of steps per subsequent win). All distributions use the blocked-curricula protocol with curriculum steps included, matching the main-text aggregate figures.

\clearpage
\subsection{Level Progression Under Blocked Curricula}
\label{app:curriculum}

\begin{figure}[h]
\centering
\includegraphics[width=\textwidth]{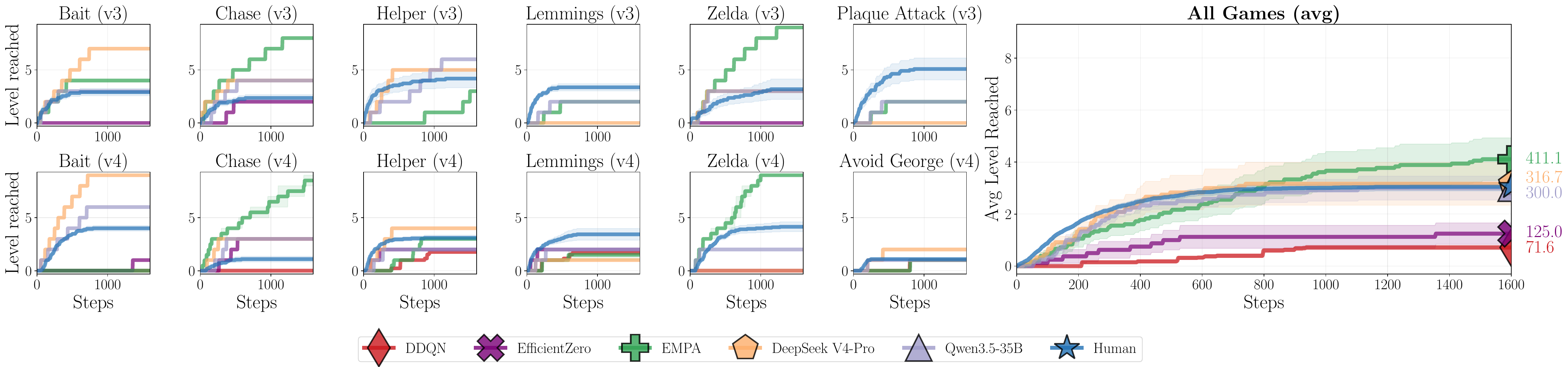}
\caption{\textbf{Level progression -- cross-paradigm.}
  Average level reached vs.\ cumulative steps under blocked curricula (2 consecutive wins to advance).
  Shaded bands show $\pm 1$ SEM across instances.
  Per-game panels (left) and pooled average (right).}
\label{fig:curriculum_cross_paradigm_supp}
\end{figure}

\begin{figure}[h]
\centering
\includegraphics[width=\textwidth]{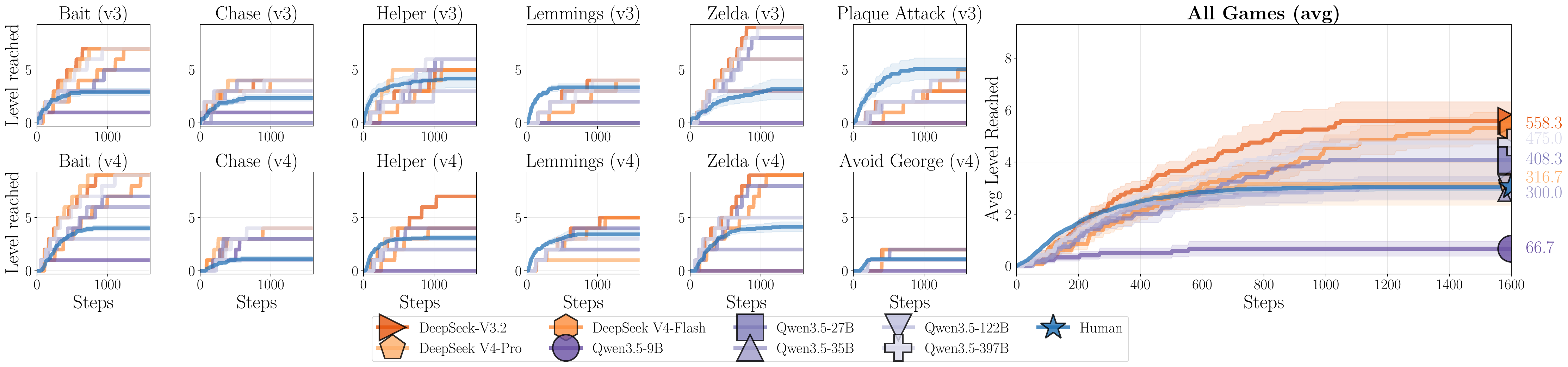}
\caption{\textbf{Level progression -- all LRMs.}
  Same format as \Cref{fig:curriculum_cross_paradigm_supp} for all eight frontier LRMs.}
\label{fig:curriculum_llms_supp}
\end{figure}

\Cref{fig:curriculum_cross_paradigm_supp,fig:curriculum_llms_supp} show the average level reached as a function of cumulative steps under the blocked-curricula protocol (2 consecutive wins required to advance). Each curve is the mean across all (game, cohort, instance) trajectories for an agent; shaded bands show $\pm 1$ SEM. Trajectories that stall on a level (failing to achieve mastery) hold their last level for the remainder of the budget, pulling the average down. This provides a complementary view to the per-level discovery and execution metrics: it captures both the rate of rule discovery and the ability to consistently execute learned strategies.

Frontier LRMs progress through levels at a rate comparable to humans, reaching an average of 4--6 levels within the 1600-step budget. The best-performing LRMs (DeepSeek~V3.2, V4-Flash, Qwen3.5-397B) closely track the human curve across the full budget. RL baselines plateau early: DDQN and EfficientZero rarely advance past level 1--2 within the same budget, reflecting their orders-of-magnitude slower learning. EMPA progresses at a rate between the RL baselines and the LRMs, consistent with its strong per-level discovery but more variable execution.

\clearpage
\subsection{Trial-by-Trial Outcome Grids}
\label{app:trial_grids}

\begin{figure}[h]
\centering
\includegraphics[width=\textwidth]{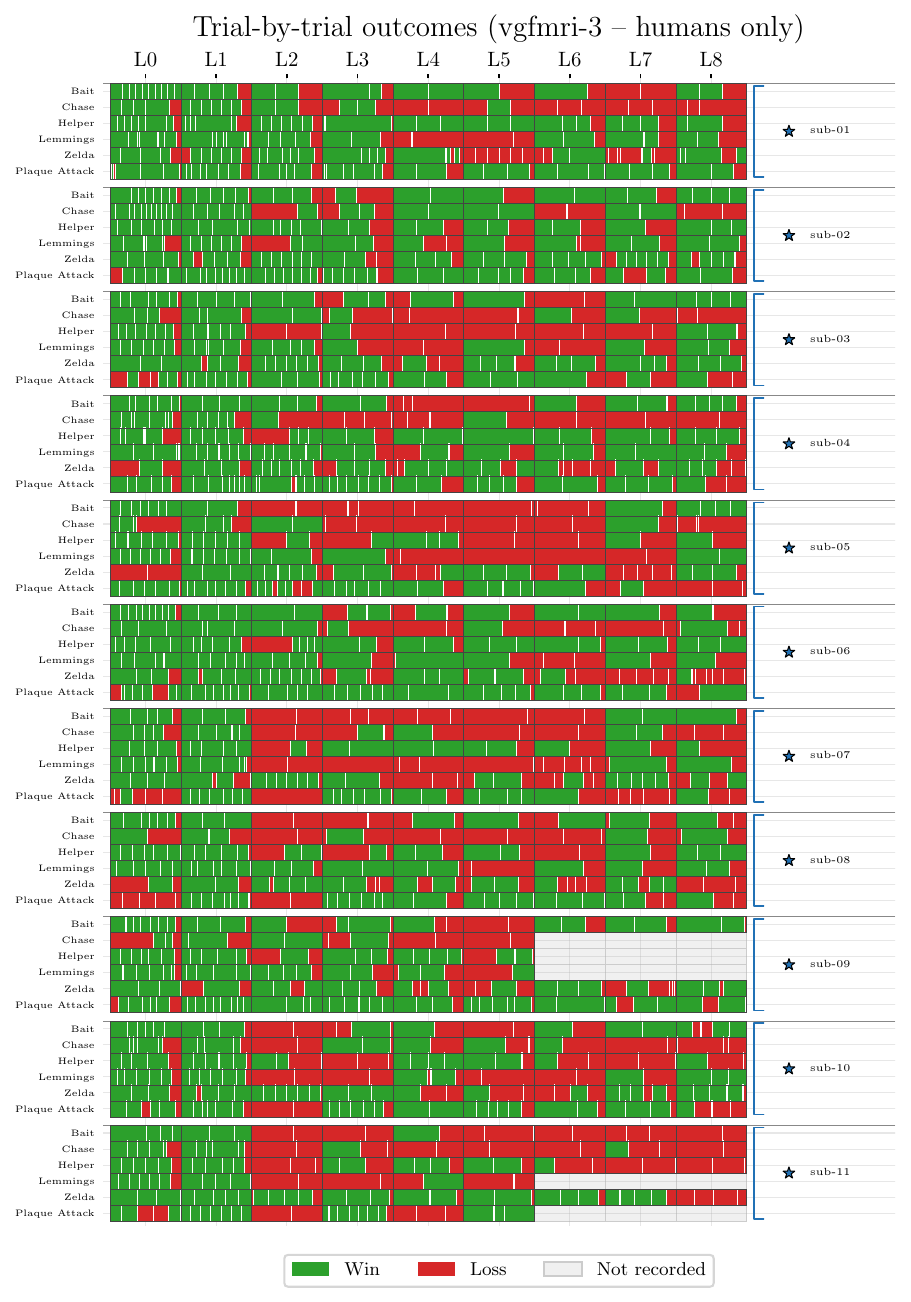}
\caption{\textbf{Human trial-by-trial outcomes -- raw progression (vgfmri3).}
  Every episode for 11 human participants (sub-01 through sub-11) on the 6 vgfmri3 games under the original fixed-budget protocol. Green = win, red = loss.}
\label{fig:trial_grid_vgfmri3_humans}
\end{figure}

\begin{figure}[h]
\centering
\includegraphics[width=\textwidth]{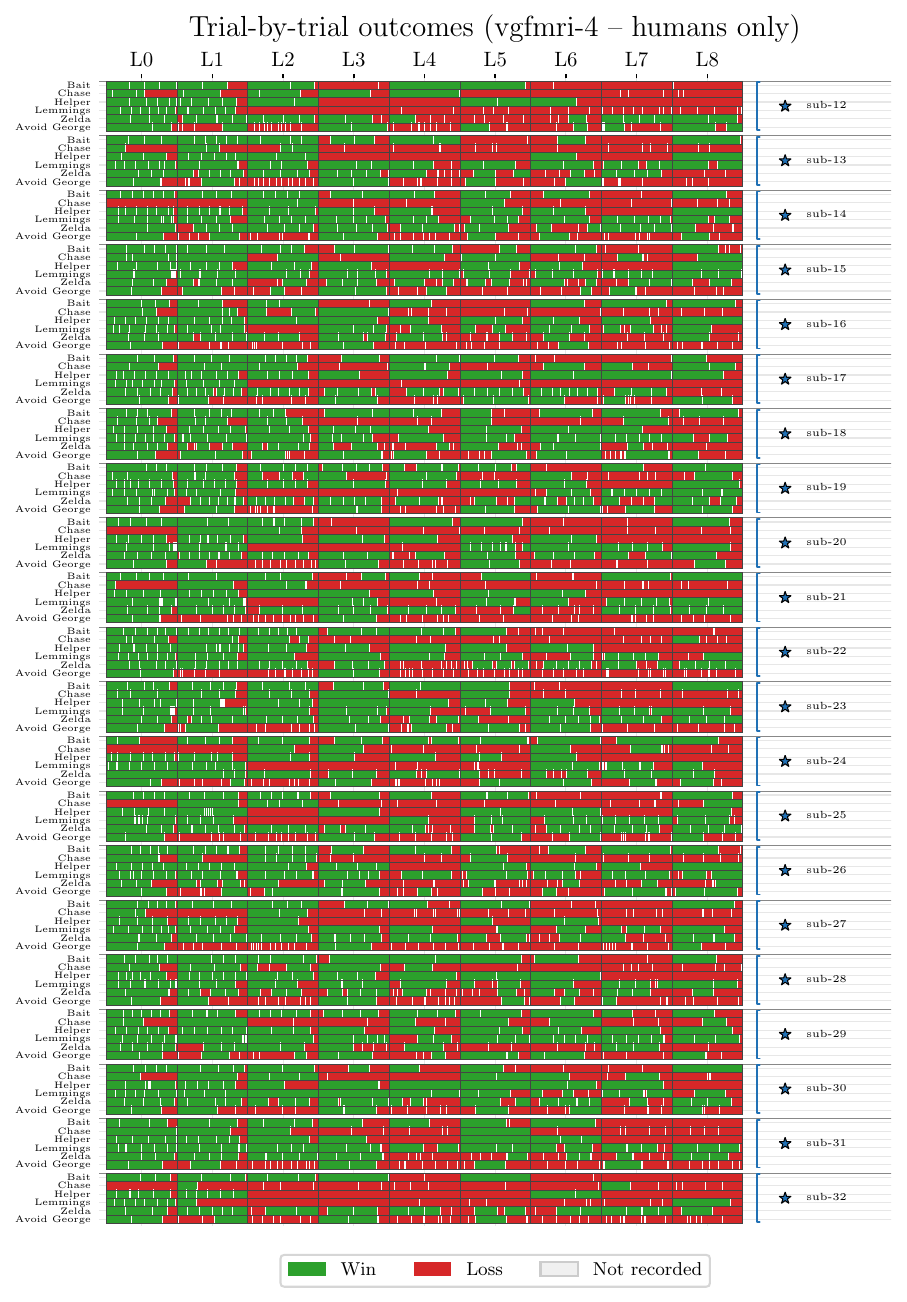}
\caption{\textbf{Human trial-by-trial outcomes -- raw progression (vgfmri4).}
  Same format as \Cref{fig:trial_grid_vgfmri3_humans} for 21 human participants (sub-12 through sub-32) on the 6 vgfmri4 games.}
\label{fig:trial_grid_vgfmri4_humans}
\end{figure}

\begin{figure}[h]
\centering
\includegraphics[width=\textwidth]{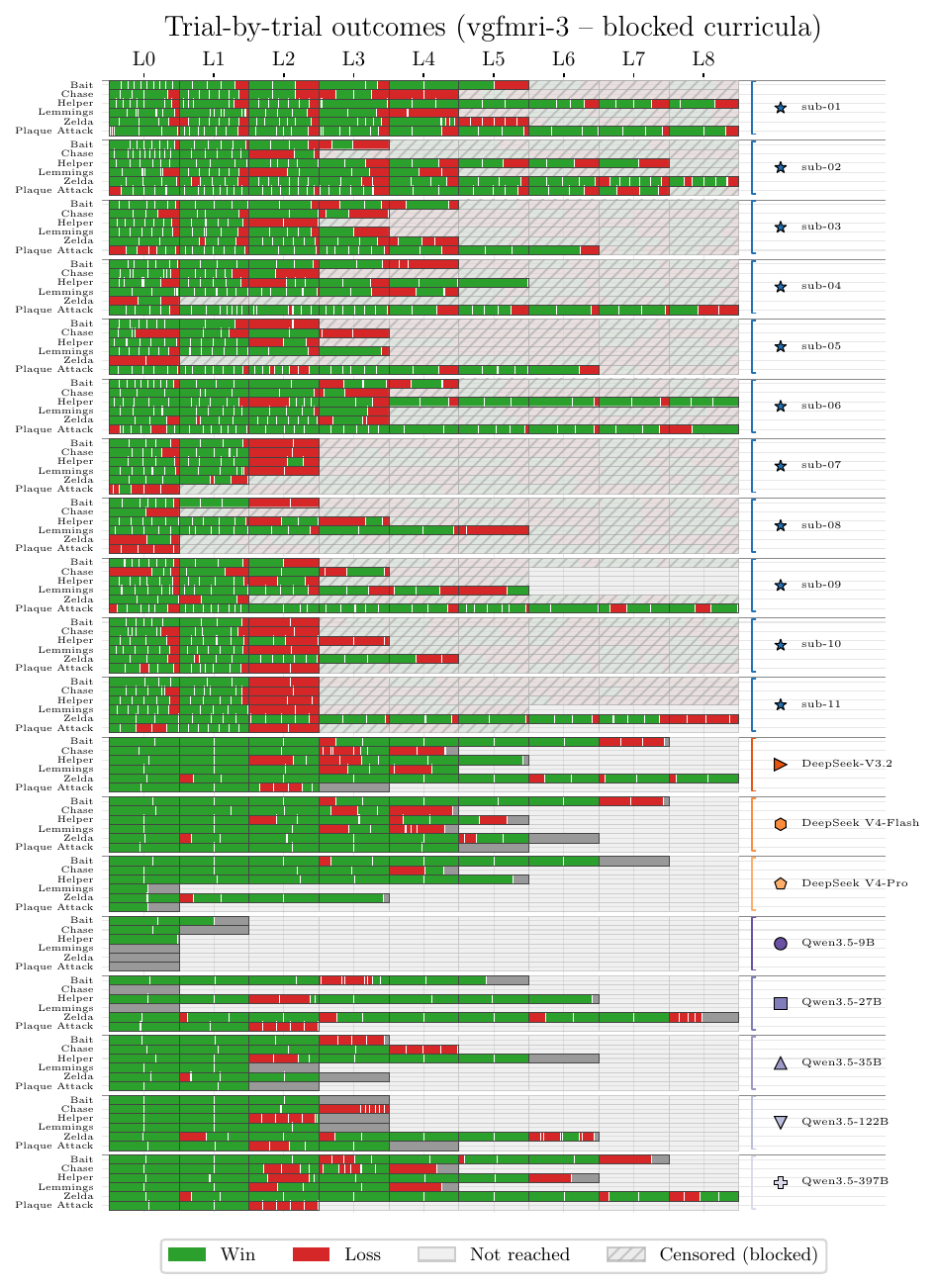}
\caption{\textbf{Trial-by-trial outcomes -- blocked curricula (vgfmri3).}
  All agents (11 humans + 8 LRMs) on the 6 vgfmri3 games. Levels after first failure to achieve 2 consecutive wins are censored (hatched grey), matching the criterion applied in all behavioral and encoding analyses.}
\label{fig:trial_grid_vgfmri3_blocked}
\end{figure}

\begin{figure}[h]
\centering
\includegraphics[width=\textwidth]{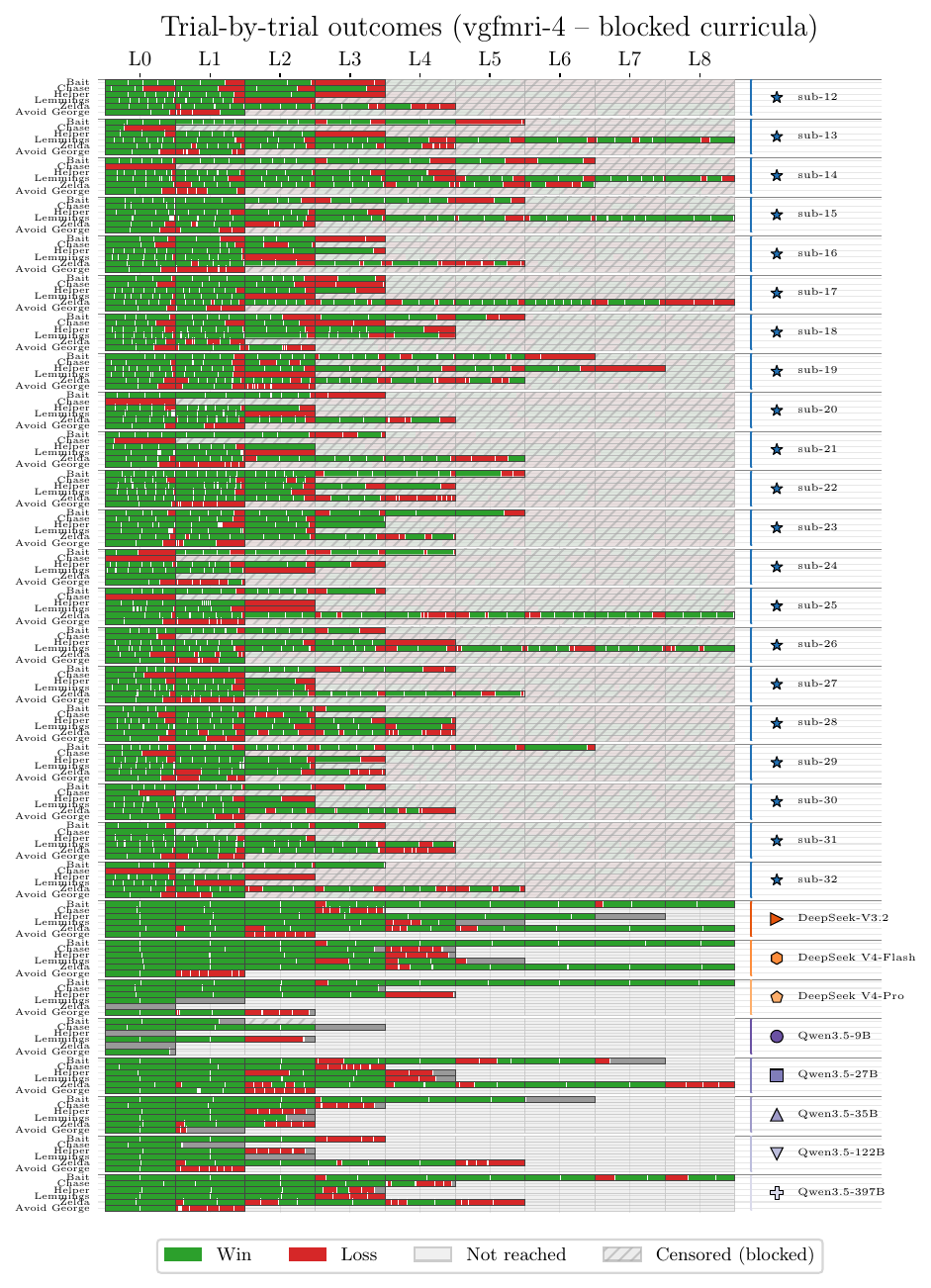}
\caption{\textbf{Trial-by-trial outcomes -- blocked curricula (vgfmri4).}
  Same format as \Cref{fig:trial_grid_vgfmri3_blocked} for 21 human participants and 8 LRMs on the 6 vgfmri4 games.}
\label{fig:trial_grid_vgfmri4_blocked}
\end{figure}

We present two views of the trial-by-trial outcome record. \Cref{fig:trial_grid_vgfmri3_humans,fig:trial_grid_vgfmri4_humans} show the raw human progression under the original fixed-budget protocol: each participant played all 9 levels of each game regardless of performance, advancing unconditionally after 60 seconds per level. \Cref{fig:trial_grid_vgfmri3_blocked,fig:trial_grid_vgfmri4_blocked} show all agents (humans and LRMs) under the blocked-curricula criterion used throughout our analyses: levels after the first failure to achieve two consecutive wins are retroactively censored (hatched grey overlay). This uniform censoring enables apples-to-apples comparison across agent types by imposing the same mastery requirement on all agents.

Each row corresponds to one game for one agent; columns are levels 0--8. Within each cell, the horizontal extent is divided into subcells proportional to the number of steps in each trial, colored green (win) or red (loss). Light gray cells indicate levels not reached; hatched cells indicate levels censored under blocked curricula.

\clearpage

\clearpage

\clearpage

\clearpage
\subsection{Action-Only Behavioral Ablation}
\label{sec:action_only_supp}

The copied-reasoning condition (\S\ref{sec:multiturn}) provides each LRM with its own reasoning trace from prior steps, enabling multi-step strategy formation. In the action-only condition, the model receives only the sequence of past observations and the actions it took---no reasoning traces appear in the conversation. This condition serves two purposes: (1) it isolates the contribution of explicit reasoning to behavioral performance, and (2) it validates that the passive-observer encoding protocol (\S\ref{sec:encoding}), which likewise presents only observations and actions without reasoning, still operates on representations from a model that can produce meaningful gameplay behavior.

\begin{figure}[h]
\centering
\includegraphics[width=\textwidth]{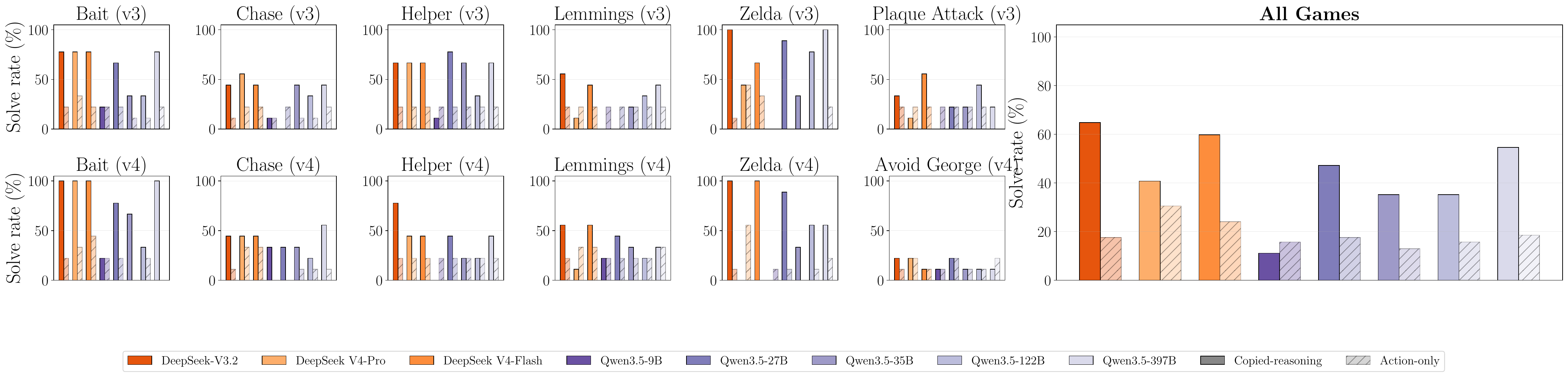}
\caption{\textbf{Effect of reasoning on solve rate.}
  Per-game solve rates for all eight frontier LRMs under the copied-reasoning (solid) and action-only (hatched) conditions.
  Removing reasoning traces from the conversation context causes a uniform collapse in performance across all models and games.}
\label{fig:rationale_compare}
\end{figure}

Figure~\ref{fig:rationale_compare} shows per-game solve rates for all eight LRMs under both conditions. Removing reasoning traces causes a dramatic and uniform collapse in performance: action-only solve rates range from 13\% to 31\% across models, compared to 11\%--65\% with copied-reasoning. The gap is largest for models that benefit most from reasoning---DeepSeek~V3.2 drops from 65\% to 18\%, and Qwen3.5-397B from 55\% to 19\%. The effect is consistent across games: no model in the action-only condition exceeds 50\% on any game-cohort pair, whereas several models solve $>$80\% of level-instances on bait and zelda with copied-reasoning.

This confirms that explicit chain-of-thought reasoning is critical for in-context game learning, not merely for articulating a strategy the model would have followed anyway. It also validates the encoding protocol: even without reasoning in context, the action-only models still solve 13\%--31\% of levels, indicating that the model's representations of game states retain meaningful structure under the same input format used for brain encoding.

\clearpage
\subsection{Reasoning Trace Case Study}
\label{app:reasoning_case_study}

The following excerpts are from a single DeepSeek~V3.2 run on \textsc{zelda} (copied-reasoning, minimal prompt, seed~0). The minimal prompt provides only the action space and response format; no game mechanics are described. The full step-by-step trace is available in the Interactive Supplementary Material (Appendix~\ref{app:interactive_supp}).

\paragraph{Self-identification (Step 0).}
The agent's first reasoning trace (12,115 characters) is dominated by the problem of determining which, if any, of the visible objects it controls:

\begin{quote}\small\itshape
We need to decide an action: up, down, left, right, action, wait. But we don't have any agent position mentioned? Possibly we are the agent? [\ldots] The objects are listed: DARKBLUE, GREEN, ORANGE. They might be targets. But we need to move to them. Without our position, we cannot plan.
\end{quote}

\paragraph{Hypothesis generation (Step 0).}
The agent systematically enumerates candidate avatar identities and reasons about what actions would be available under each hypothesis:

\begin{quote}\small\itshape
If GREEN is agent, moving right to (9,2) is possible? [\ldots] But we don't know if GREEN is agent. Maybe DARKBLUE is agent? Then at (6,9). Then we could move.
\end{quote}

\paragraph{Causal verification (Step 1).}
After choosing ``up,'' the agent observes the result and confirms its hypothesis:

\begin{quote}\small\itshape
From previous step, I chose ``up'' and now DARKBLUE is at (6,8). Previously it was at (6,9). So indeed DARKBLUE moved up. That confirms that DARKBLUE is the agent/player character.
\end{quote}

\paragraph{Spatial planning (Step 1).}
Without any instruction to optimize trajectories, the agent computes distances and orders subgoals:

\begin{quote}\small\itshape
We might want to collect GREEN first because it's closer? Compare distances: To GREEN: Manhattan distance = $|6{-}8| + |8{-}2| = 2{+}6{=}8$. To ORANGE: $|6{-}17| + |8{-}4| = 11{+}4{=}15$. So GREEN is closer. So likely we should go to GREEN first.
\end{quote}

\paragraph{Mode switching (Steps 17--21).}
After establishing a plan, the agent transitions to terse execution-only traces:

\begin{quote}\small\itshape
We are at (12,4). ORANGE at (17,4). We need to move right 5 more steps. Continue moving right. Next move: ``right'' to (13,4). Action: right
\end{quote}

\paragraph{Theory consolidation (Step 33, after first win).}
The agent synthesizes a causal theory from its accumulated experience:

\begin{quote}\small\itshape
The last step was moving left onto GREEN from (9,2) to (8,2). Then the trial ended with outcome won, score 5. So maybe moving onto GREEN with ORANGE in inventory completed the level? [\ldots] So the goal is likely to pick up ORANGE and deliver it to the GREEN location.
\end{quote}

\paragraph{Conflict resolution (Step 71).}
When the avatar dies from touching a new object (GOLD), the agent reconciles expectations with the surprising outcome:

\begin{quote}\small\itshape
We moved down onto GOLD at (3,10) and then the trial ended with outcome died, score $-1$. So moving onto GOLD caused death. That suggests GOLD is dangerous [\ldots] But the object is called GOLD, which might be a treasure, but it killed us. [\ldots] Maybe GOLD is an enemy that moves and kills us if we touch it?
\end{quote}

\paragraph{Theory simplification (Step 34, start of second attempt).}
Applying Occam's razor, the agent prunes its plan to the essentials:

\begin{quote}\small\itshape
We already know a good path: from (6,9) go to ORANGE at (17,4) first, collect it, then go to GREEN at (8,2). [\ldots] Let's plan an efficient path.
\end{quote}

\clearpage
\subsection{Per-Game Capability and Discovery Efficiency}
\label{app:capability_table}

Table~\ref{tab:capability} reports per-game solve rates and discovery efficiency for all agents. Each cell shows the fraction of (instance, level) pairs on which the agent achieved at least one win, followed by the median number of cumulative steps to first win with interquartile range. Discovery time is computed as the sum of episode steps up to and including the first winning episode on each level; levels never solved contribute to the solve rate denominator but not to the step statistics.

The best LRMs (DeepSeek~V3.2, V4-Pro) achieve solve rates within 10 percentage points of humans on most games while discovering solutions in a comparable number of steps. RL baselines that do solve levels typically require one to two orders of magnitude more steps, with DDQN showing particularly high variance (e.g., median 23,940 steps on zelda). Among the Qwen family, capability scales with model size: Qwen3.5-9B solves only 50\% of level-instances on average, while Qwen3.5-397B reaches 80\%. The DeepSeek models are more uniformly capable across games, with V3.2 achieving the highest overall solve rate (87\%).

\begin{table}[h]
\centering
\caption{\textbf{Per-game capability and discovery efficiency.} Each cell shows level-solve rate / median discovery steps (Q1--Q3). Discovery = cumulative steps to first win on a level. \textsc{Avg} = macro-average across games. LRM results use the copied-reasoning + elaborate condition.}
\label{tab:capability}
\resizebox{\textwidth}{!}{%
\begin{tabular}{lrrrrrrrr}
\toprule
Agent & \textsc{bait} & \textsc{chase} & \textsc{helper} & \textsc{lemmings} & \textsc{zelda} & \textsc{plaqueattack} & \textsc{avoidgeorge} & \textsc{Avg} \\
\midrule
Human & 78\% / 75 (50--112) & 54\% / 124 (80--188) & 79\% / 71 (32--106) & 78\% / 46 (26--80) & 93\% / 62 (44--86) & 86\% / 47 (25--78) & 63\% / 86 (68--110) & 76\% / 73 \\
\midrule
EMPA 2.0 & 6\% / 127 & 77\% / 455 (142--586) & 72\% / 360 (221--947) & 22\% / 298 (275--310) & 13\% / 197 (197--284) & 100\% / 351 (210--384) & 89\% / 400 (400--406) & 54\% / 312 \\
EfficientZero & 50\% / 108 (16--1,608) & 100\% / 145 (65--221) & 75\% / 66 (32--232) & 50\% / 31 (25--397) & 82\% / 386 (71--24,648) & 100\% / 68 (38--114) & 100\% / 408 (100--722) & 80\% / 173 \\
DDQN & 44\% / 1,474 (1,473--5,596) & 35\% / 1,580 (1,477--8,748) & 48\% / 261 (237--5,835) & 54\% / 215 (127--2,904) & 38\% / 23,940 (11,403--38,240) & -- & 63\% / 471 (398--2,188) & 47\% / 4,657 \\
\midrule
Qwen3.5-9B & 80\% / 42 (34--49) & 67\% / 46 (29--90) & 50\% / 51 & 50\% / 67 (64--70) & 0\% & 0\% & 100\% / 100 & 50\% / 61 \\
Qwen3.5-27B & 93\% / 56 (36--119) & 60\% / 35 (29--50) & 92\% / 60 (55--117) & 67\% / 78 (70--94) & 89\% / 54 (40--81) & 67\% / 84 (84--84) & 67\% / 100 (100--100) & 76\% / 67 \\
Qwen3.5-35B-A3B & 82\% / 55 (38--68) & 78\% / 68 (39--86) & 80\% / 72 (66--80) & 83\% / 79 (72--84) & 86\% / 39 (36--48) & 67\% / 104 (96--112) & 50\% / 100 & 75\% / 74 \\
Qwen3.5-122B-A10B & 75\% / 34 (28--46) & 83\% / 62 (19--73) & 71\% / 51 (51--78) & 71\% / 79 (72--84) & 92\% / 51 (31--62) & 80\% / 108 (92--194) & 50\% / 100 & 75\% / 69 \\
Qwen3.5-397B-A17B & 94\% / 70 (36--98) & 90\% / 82 (37--116) & 91\% / 62 (54--73) & 78\% / 84 (76--120) & 93\% / 52 (48--68) & 67\% / 100 (94--106) & 50\% / 100 & 80\% / 78 \\
\midrule
DeepSeek-V3.2 & 94\% / 44 (31--58) & 89\% / 68 (29--153) & 93\% / 63 (54--108) & 91\% / 90 (80--226) & 100\% / 49 (38--54) & 75\% / 79 (75--94) & 67\% / 102 (101--102) & 87\% / 71 \\
DeepSeek-V4-Flash & 94\% / 76 (61--106) & 80\% / 86 (46--124) & 91\% / 120 (99--149) & 82\% / 168 (158--185) & 91\% / 54 (49--68) & 83\% / 142 (121--170) & 50\% / 200 & 82\% / 121 \\
DeepSeek-V4-Pro & 94\% / 48 (33--59) & 100\% / 40 (30--124) & 91\% / 52 (46--54) & 67\% / 70 (66--75) & 80\% / 54 (46--55) & 100\% / 85 & 67\% / 108 (104--111) & 85\% / 65 \\
\bottomrule
\end{tabular}%
}
\end{table}



\subsection{Methodological Refinements Over Tomov et al.\ 2023}
\label{app:tomov_fixes}

Our work builds on the VGDL-fMRI dataset and analysis framework of \citet{tomov2023neural}. In the course of extending their pipeline to support the larger model set and the encoding analyses in this paper, we identified and corrected several issues in the original codebase and methodology. We document these here for transparency and to support future work using this dataset.

\subsubsection{DDQN implementation corrections}
\label{app:ddqn_bugs}

The DDQN agent used in \citet{tomov2023neural} contained seven implementation errors that collectively degraded training quality and reproducibility. We enumerate them below with references to the original source file at commit \texttt{366fc03} (the latest commit at the time of writing): \url{https://github.com/tomov/RC_RL/blob/366fc03/player.py}.

\begin{enumerate}
\item \textbf{Broken action selection} (\href{https://github.com/tomov/RC_RL/blob/366fc03/player.py#L155}{\texttt{player.py:155}}). Random exploration actions were hardcoded to choose between actions 0 and 1 only (\texttt{np.random.choice([0, 1])}), ignoring the actual action-space size \texttt{self.n\_actions}. In games with 6 available actions (4 direction, action button, and NO\_OP), the agent could never explore four of them.

\item \textbf{Incorrect state representation} (\href{https://github.com/tomov/RC_RL/blob/366fc03/player.py#L266}{\texttt{player.py:266}}, \href{https://github.com/tomov/RC_RL/blob/366fc03/player.py#L316}{\texttt{316}}). The state was computed as a frame difference (\texttt{current\_screen - last\_screen}) rather than using stacked frames or raw frames. Frame differences are unreliable and discard absolute position information, degrading the quality of the state signal available to the network.

\item \textbf{Undiscriminating reward clipping} (\href{https://github.com/tomov/RC_RL/blob/366fc03/player.py#L306}{\texttt{player.py:306}}). All rewards were clamped to $[-1, 1]$, collapsing the reward structure. Different reward magnitudes---which carry distinct semantic meaning in VGDL games---became indistinguishable.

\item \textbf{Inconsistent model loading} (\href{https://github.com/tomov/RC_RL/blob/366fc03/player.py#L98-L99}{\texttt{player.py:98--99}}). The checkpoint-loading code loaded the same saved weights into the target network twice in consecutive lines, never updating the policy network. As a result, the policy network retained its random initialization after loading a checkpoint.

\item \textbf{Flawed model-update logic} (\href{https://github.com/tomov/RC_RL/blob/366fc03/player.py#L140}{\texttt{player.py:140}}). The save condition used \texttt{or} instead of \texttt{and}: \texttt{if self.episode\_reward > self.best\_reward or self.steps \% 50000}. This saved the model every 50k steps regardless of performance, overwriting good checkpoints with potentially worse ones.

\item \textbf{Wrong variable reference for target updates} (\href{https://github.com/tomov/RC_RL/blob/366fc03/player.py#L136}{\texttt{player.py:136}}). Target-network updates were gated on total cumulative \texttt{self.steps} rather than an episode-specific counter, causing target updates at inconsistent intervals and breaking learning stability.

\item \textbf{Seed-setting bug} (\href{https://github.com/tomov/RC_RL/blob/366fc03/player.py#L251}{\texttt{player.py:251}}). The seed was assigned to the function object rather than called: \texttt{torch.manual\_seed = (self.config.random\_seed)} instead of \texttt{torch.manual\_seed(self.config.random\_seed)}. The random seed was never actually set, eliminating reproducibility.
\end{enumerate}

Beyond these code-level fixes, we made two methodological changes to the DDQN training protocol. First, we replaced the original sequential overfitting regime---250k on-policy gradient updates per level, looping through all 9 levels for 100 epochs---with a curriculum-based protocol that trains a single agent across all 9 levels with a fixed budget of 100k gradient updates per level (Appendix~\ref{app:ddqn}). Second, in contrast to the single shared configuration used across all games in \citet{tomov2023neural}, we conducted an extensive per-game Hyperband sweep (256 configurations $\times$ 4 successive-halving stages per game) over learning rate, discount, batch size, replay-buffer size, target-update frequency, $\epsilon$-decay schedule, gradient clip, and frame-stack size; full ranges are listed in Appendix~\ref{app:ddqn}. These changes raise the performance bar for the DDQN baseline and ensure that any remaining gap between DDQN and LRMs reflects a genuine capability difference rather than an undertrained baseline.

\subsubsection{State representation}
\label{app:state_repr_fixes}

The original pipeline had two compounding state-representation issues.

\paragraph{RGB pixels instead of sprite-type channels.}
The DDQN agent received direct RGB pixel renderings of the game screen as input (\href{https://github.com/tomov/RC_RL/blob/366fc03/player.py#L75-L79}{\texttt{player.py:75--79}}). In VGDL, color is the sole carrier of object identity: each sprite type is rendered as a distinct color, and participants (human or model) must learn which colors correspond to which interaction rules. Feeding raw RGB conflates perceptual decoding (segmenting colors) with rule learning, forcing the network to jointly solve both problems. We replaced this with a per-sprite-type channel representation: each sprite type is assigned a dedicated binary channel, and the observation tensor has shape $(C \times H \times W)$ where $C$ is the number of sprite types. This isolates the rule-learning problem from the perceptual one and provides a semantically meaningful input across levels and games.

\paragraph{Aspect-ratio distortion across levels.}
All level layouts were stretched and rescaled to a single fixed spatial resolution (\href{https://github.com/tomov/RC_RL/blob/366fc03/player.py#L54-L58}{\texttt{player.py:54--58}}), regardless of the original grid dimensions. VGDL levels within a game---and especially across games---vary in width and height, so this rescaling produced variable tile sizes and aspect ratios. The network had to learn that the ``same'' sprite occupies differently shaped regions across levels, adding a spurious source of variability. We instead preserve the native grid dimensions ($H \times W$ matching the level layout) so that each tile occupies exactly one spatial position in the input tensor, eliminating aspect-ratio distortion entirely.

\subsubsection{Random-initialization controls}
\label{app:random_init_fixes}

\begin{figure}[h]
    \centering
    \includegraphics[width=\textwidth]{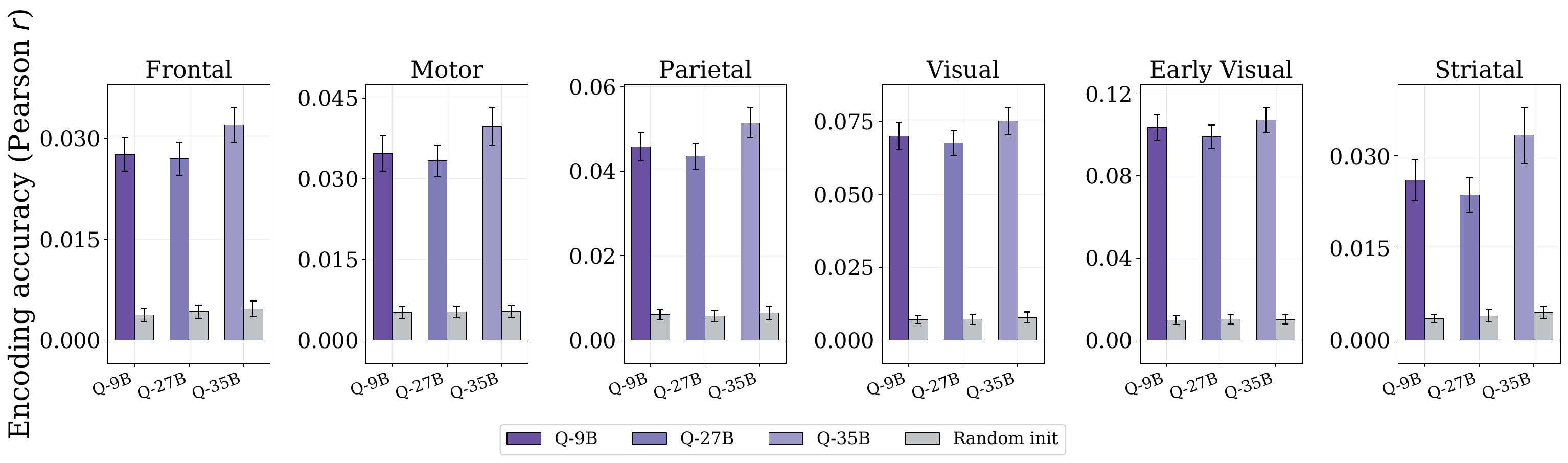}
    \caption{%
        \textbf{LRM encoding accuracy is not reproduced by
        randomly-initialised weights.}
        For each of three Qwen3.5 sizes (9B, 27B, 35B-A3B), we
        compare encoding accuracy of the trained model (coloured bars)
        to a randomly-initialised version with matched architecture
        (grey bars). Bars show per-subject best-layer Pearson $r$,
        $\pm$ SEM across $n=21$ subjects, broken out by region group.
        Trained models substantially outperform their random-init
        counterparts in every region group across all three sizes,
        confirming that the encoding signal depends on training rather
        than on architectural inductive biases of the transformer
        forward pass alone.%
    }
    \label{supp:random_init}
\end{figure}

\citet{tomov2023neural} reported significant encoding differences between DDQN and EMPA but did not include random-initialization controls. Recent work has shown that untrained neural networks with matching architecture can approach the encoding performance of trained models on video-game stimuli \citep{paugam2025brittle}, raising the question of whether trained-vs-shuffled permutation tests provide a meaningful baseline. We include same-architecture random-initialization controls for the Qwen~3.5 family (9B, 27B, 35B-a3B) and report the trained-vs-random gap explicitly (Section~\ref{sec:encoding_headline}, Figure~\ref{supp:random_init}). The $6.4$--$6.9\times$ gap confirms that LRM encoding accuracy reflects learned representations rather than architectural priors.

\subsection{Lessons for Future Experimental Design}
\label{app:lessons}

The VGDL-fMRI dataset of \citet{tomov2023neural} was designed for ecological validity, not factorial control. While this makes it a powerful ``in the wild'' testbed, several structural properties limit the precision of behavioral and neural comparisons. We document these here as guidance for future data collection efforts in this paradigm.

\subsubsection{Noise ceiling estimation is intractable on this dataset}
\label{app:noise_ceiling}

A standard approach to estimating the reliability ceiling for brain encoding is to measure between-trial or between-session consistency for repeated presentations of identical stimuli \citep{lage2019methods}. This approach is inapplicable here because every participant's trajectory is unique: participants learn on-policy, so each person's action sequence---and therefore the sequence of observations, rewards, and BOLD-eliciting events---diverges from every other participant's within the first few steps of gameplay.

\begin{figure}[h]
\centering
\includegraphics[width=0.85\textwidth]{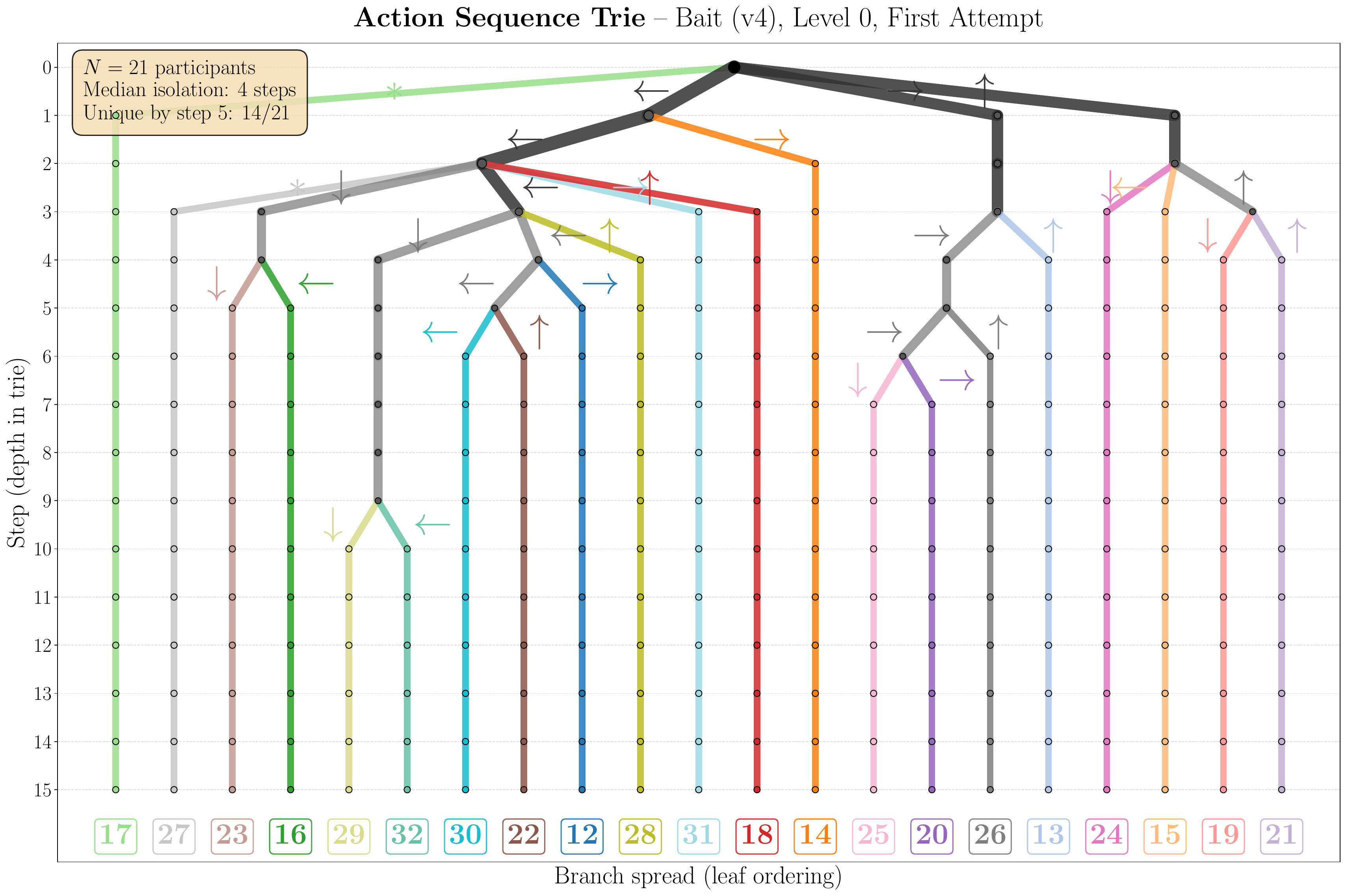}
\caption{\textbf{Action-sequence prefix trie for Bait (vgfmri4), Level~0, first attempt.}
  Each branch traces one participant's action sequence from the shared root.
  Arrow symbols denote actions ($\uparrow\downarrow\leftarrow\rightarrow$: movement; $\ast$: action; $\circ$: wait).
  Edge thickness and color encode the number of participants sharing that prefix (thick dark = many; thin colored = individual).
  Leaf labels show participant IDs.
  The cohort fans out almost immediately: median isolation depth is 4 steps, and 14/21 participants are fully unique by step~5.}
\label{fig:action_trie}
\end{figure}

\begin{figure}[h]
\centering
\includegraphics[width=\textwidth]{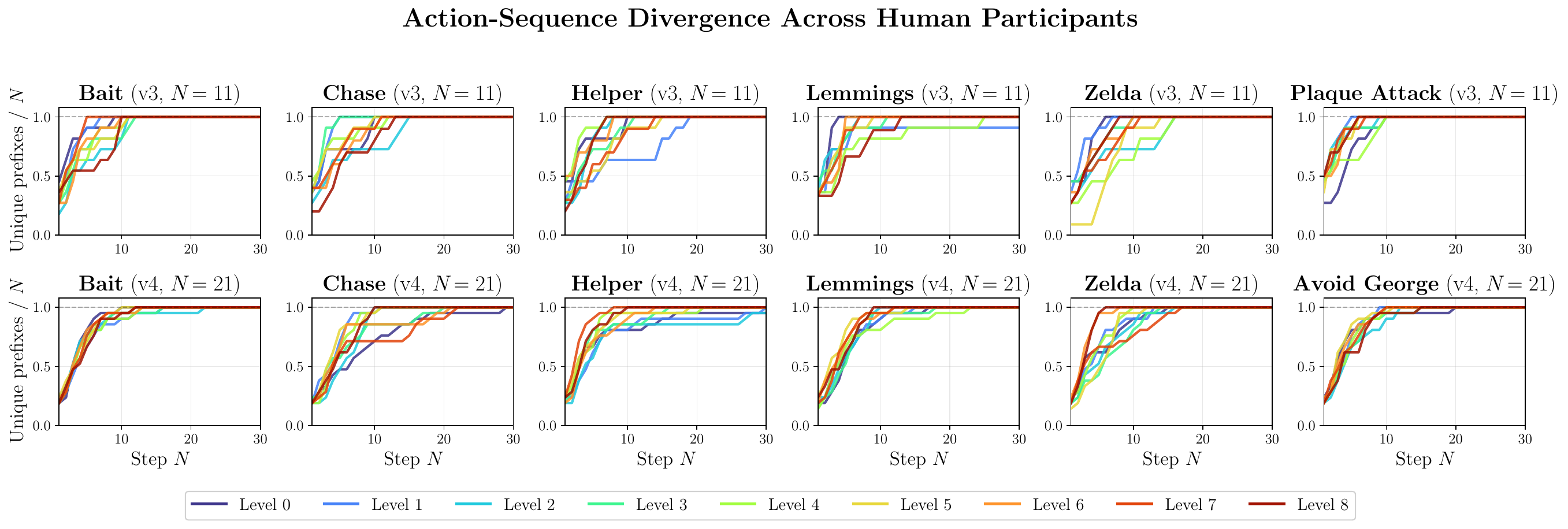}
\caption{\textbf{Action-sequence divergence across all 12 games.}
  Each curve shows the fraction of unique action prefixes (unique prefixes / $N$ participants) as a function of step depth for one level (color = level index).
  Top row: vgfmri3 cohort ($N=11$); bottom row: vgfmri4 ($N=21$).
  Across nearly all games and levels, the cohort approaches complete divergence within 10--15 steps, confirming that no two participants follow the same trajectory for long enough to support traditional noise-ceiling estimation.}
\label{fig:branch_divergence}
\end{figure}

We quantify this divergence directly. For each (game, level, first attempt), we build a prefix trie of all participants' action sequences and measure how many steps elapse before each participant's prefix becomes unique (``isolation depth''). Figure~\ref{fig:action_trie} shows a representative example: on Bait (vgfmri4, Level~0), the median isolation depth is 4 steps, and 14 of 21 participants are fully unique by step~5. Figure~\ref{fig:branch_divergence} extends this analysis across all 12 games and 9 levels, plotting the fraction of unique action prefixes as a function of step depth. Across nearly every (game, level) combination, the cohort reaches near-complete divergence within 10--15 steps.

This rapid divergence means there are essentially no repeated stimulus sequences across participants, and within-participant repetitions (re-attempts on the same level) differ because the participant's knowledge state has changed between attempts. Traditional split-half or between-subject reliability estimates therefore have no valid basis in this data.

We recommend two design features for future experiments:
\begin{itemize}
\item \textbf{Repeat-trial blocks.} After achieving mastery on a level (e.g., $N$ consecutive wins), require $K$ additional replay trials. At this point the participant already knows the rules, so behavioral variance is dominated by execution noise rather than learning. These repeated post-mastery trials provide a within-subject reliability estimate suitable for computing a noise ceiling.
\item \textbf{Passive replay viewing.} After completing gameplay, present participants with a video replay of their own (or another participant's) gameplay as a passive viewing task. Correlating active-play BOLD with passive-viewing BOLD for matched stimuli gives a cross-modality reliability ceiling that separates stimulus-driven signal from action-planning signal.
\item \textbf{Corridor-and-junction level design.} Design levels so that execution between decision points is spatially linear: long corridors or single-path segments where the only reasonable action is to walk forward. All participants traversing the same corridor produce identical (or near-identical) action sequences, yielding repeated stimulus epochs suitable for reliability estimation. Meaningful cognitive variance is then concentrated at junction points where the participant must choose between semantically distinct alternatives (e.g., collect the key or avoid the enemy first). This separates rote navigation from genuine decision-making and creates natural alignment anchors across participants without sacrificing the validity of the overall task.
\end{itemize}

\subsubsection{Confounded cognitive demands}
\label{app:confounded_demands}

Each VGDL game requires multiple cognitive capacities simultaneously, and no game in the current battery isolates a single axis. When a model predicts brain activity well on a given game, we cannot attribute the alignment to any one cognitive component. We enumerate the entangled axes below, with examples from the vgfmri3/4 game set.

\paragraph{Spatial and perceptual.}
Spatial planning and pathfinding (bait, helper); spatial memory for which tiles are dangerous after a single interaction (zelda); visual tracking under clutter (lemmings, plaque attack).

\paragraph{Temporal and reactive.}
Agility and reaction time under fast-moving NPCs (chase, plaque attack); time pressure vs.\ deliberation tradeoff imposed by the 60-second window; sequential timing of when to act (plaque attack shooting, bait pushing).

\paragraph{Rule learning and induction.}
Interaction-rule discovery (what happens when the avatar touches a given color?); conditional rules (e.g., \texttt{killIfOtherHasMore}: a sprite is only removed if the avatar holds a required resource); transformation chains (object A becomes B becomes C across levels).

\paragraph{Higher-order reasoning.}
Tool use and composition (pushing a block onto a target to create a bridge); subgoal decomposition (collect key, then unlock door, then reach goal); goal composition with ordering constraints (bait: items must be collected in the correct sequence); theory of mind and agent modeling (helper: predicting the cooperative NPC's intent; lemmings: predicting autonomous NPC trajectories; chase: modeling pursuit strategies; avoid george: modeling adversarial agent behavior).

\paragraph{Meta-cognitive.}
Hypothesis formation and revision; exploration vs.\ exploitation decisions; error attribution (which sprite caused the avatar's death?); transfer of learned rules across levels within the same game.

\paragraph{Missing dimension: adversarial rule changes.}
No game in the current battery changes a rule mid-session. All interaction rules are stable across levels within a game; later levels add new rules but never contradict previously learned ones. A future battery should include games with concept overwriting---where a previously safe sprite becomes lethal, a collection rule reverses, or interaction effects swap between levels---to test belief revision under contradiction, a core cognitive capacity that the current games never probe.

\paragraph{Recommendations.}
Since think-aloud protocols are likely infeasible during fMRI scanning (vocalization artifacts, head motion), we recommend a structured post-game cognitive survey administered immediately after each scanning session. The survey should capture: which rules the participant believes they discovered and on which level; which interactions surprised them; their subjective strategy (exploration-first vs.\ goal-directed); confidence ratings per rule; and self-reported moments of insight. This provides a partial substitute for think-aloud data that is compatible with fMRI acquisition constraints.

For future game batteries, we recommend a factorial design: orthogonal game pairs that isolate one cognitive axis per comparison (e.g., two games identical except that one requires spatial planning and the other does not), and ``lesion'' games that specifically remove one capacity (e.g., all rules displayed from the start, removing rule induction and isolating planning). This would enable the kind of component-level attribution that the current ecologically valid but scientifically opaque battery does not support.

\subsubsection{Curriculum protocol limitations}
\label{app:curriculum_limitations}

The original experimental protocol of \citet{tomov2023neural} uses a fixed-budget curriculum: each participant plays each level for 60 seconds and is advanced to the next level unconditionally, regardless of whether they solved the current one. This design introduces two interrelated confounds.

\paragraph{Advancement without solving.}
A substantial fraction of (subject, level) pairs are advanced to the next level without the participant ever achieving a win. We quantify this per game in Table~\ref{tab:unsolved_advancement}. Across the full dataset, 24.4\% of all level-instances end without a single win, with large per-game variation (7.4\% on zelda to 46.3\% on chase). This means that for these (subject, level) pairs, the ``discovery'' metric is undefined, the ``execution'' metric is vacuous, and any cross-subject behavioral comparison on that level conflates genuine learning differences with whether a participant happened to encounter the right interaction within the time window.

\begin{table}[h]
\centering
\caption{\textbf{Fraction of (subject, level) pairs advanced without solving.}
  For each game, the percentage of level-instances where the participant was
  advanced to the next level without ever winning the current one.
  Computed across all 32 participants and 9 levels per game.}
\label{tab:unsolved_advancement}
\begin{tabular}{lrrr}
\toprule
Game & Total pairs & No win & \% advanced without solving \\
\midrule
\textsc{avoidgeorge} & 189 & 70 & 37.0\% \\
\textsc{bait} & 312 & 69 & 22.1\% \\
\textsc{chase} & 309 & 143 & 46.3\% \\
\textsc{helper} & 309 & 66 & 21.4\% \\
\textsc{lemmings} & 306 & 66 & 21.6\% \\
\textsc{plaqueattack} & 120 & 17 & 14.2\% \\
\textsc{zelda} & 312 & 23 & 7.4\% \\
\midrule
\textbf{All games} & 1857 & 454 & 24.4\% \\
\bottomrule
\end{tabular}
\end{table}

\paragraph{Non-monotonic level difficulty.}
The curriculum is designed so that later levels introduce additional interaction rules, ostensibly increasing difficulty. However, some levels are objectively easier than their predecessors---for instance, a level may have a simpler spatial layout or fewer hazards despite its later position in the sequence. When a participant fails level $k$ but succeeds on level $k{+}1$, it is ambiguous whether this reflects learning transfer from level $k$ or simply that level $k{+}1$ is intrinsically easier. This confound is inseparable from the advancement-without-solving issue: a participant pushed past an unsolved level who then wins the next level may appear to have ``learned'' when in fact the task became easier.

\paragraph{Implications for our analyses.}
For the behavioral analyses in this paper, we impose a blocked-curricula criterion retroactively: agents (human or model) must achieve two consecutive wins before advancing, and all subsequent levels are censored if this criterion is not met (\S\ref{sec:multiturn}). This eliminates the advancement-without-solving confound for cross-agent comparisons but reduces the effective sample size per level. For the brain encoding analyses, which use the full human gameplay trajectories as recorded, the confound remains: BOLD signal from levels the participant never solved reflects a mixture of exploration, confusion, and partial learning that may differ qualitatively from signal during successful gameplay. Future experiments should adopt a blocked curriculum with explicit mastery criteria to ensure that all analyzed gameplay reflects genuine engagement with the level's demands.

%
%
%
%
%
%

\section{Supplementary Results --- Brain Encoding}
\label{supp:encoding}

This section provides extended analyses underlying the main-text encoding
results. We first show per-ROI breakdowns of encoding accuracy
(Sec.~\ref{supp:per_roi}), then report a battery of control and robustness
analyses (Sec.~\ref{supp:controls}), and finally summarise layer-by-region
structure and the surface-projected significance maps across all seven
LRMs (Sec.~\ref{supp:heatmaps_surfaces}).


\subsection{Per-ROI encoding accuracy}
\label{supp:per_roi}

\begin{figure}[h]
    \centering
    \includegraphics[width=\textwidth]{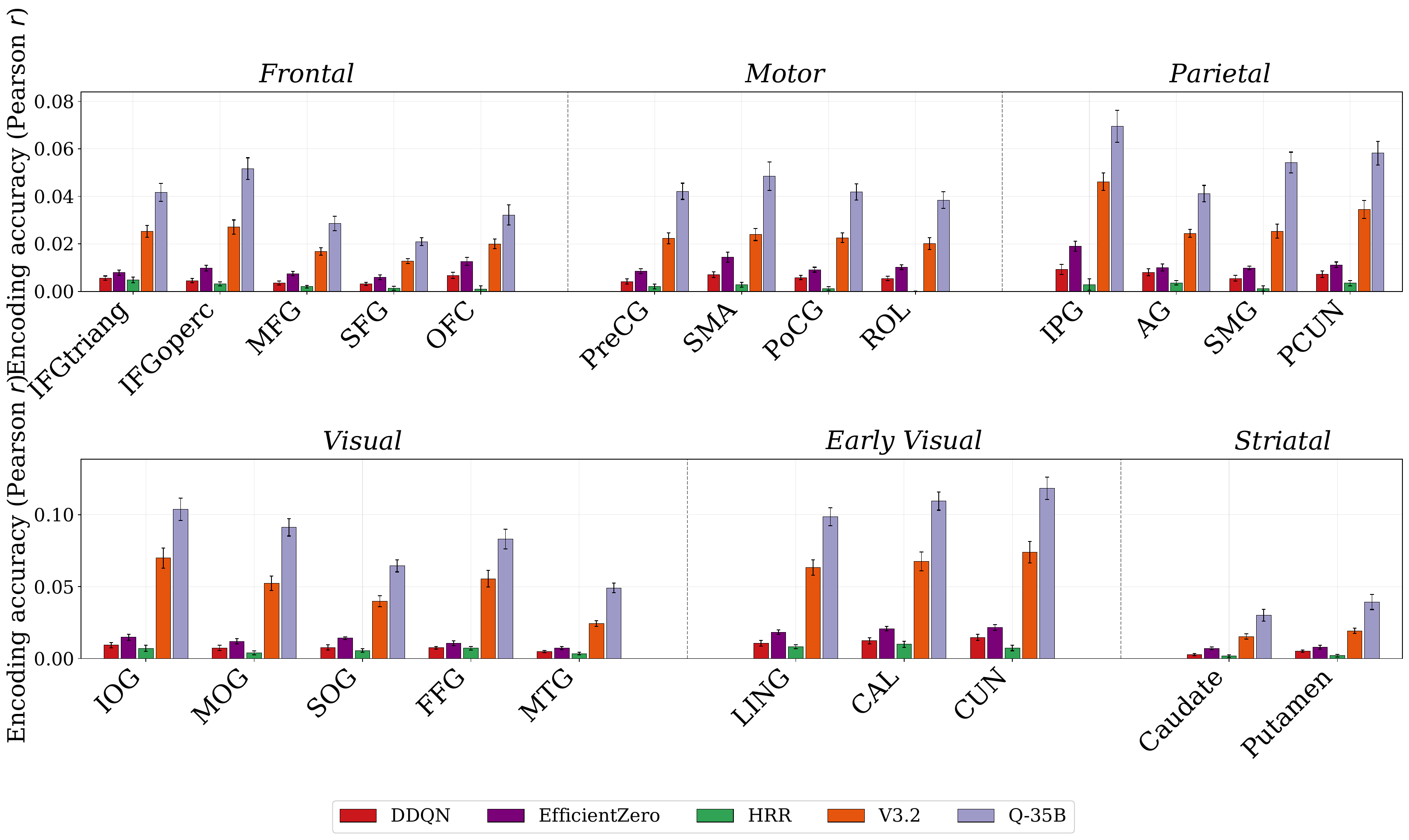}
    \caption{%
        \textbf{Per-ROI encoding accuracy for RL baselines.}
        Bars show mean encoding accuracy (per-subject best-layer Pearson
        $r$, $\pm$ SEM across $n=21$ subjects, band~=~main) for each of
        the three RL baselines (DDQN, EfficientZero, HRR) at every
        anatomical ROI in our analysis. Hatched bars represent worst-
        and median-layer encoding for multi-layer models; HRR has a
        single representation layer. Encoding accuracy is uniformly low
        ($r < 0.01$) and only weakly differentiates between regions ---
        consistent with these compact, action-focused representations
        carrying limited information about the rich sensory and
        cognitive content of human gameplay.%
    }
    \label{supp:rois_baseline}
\end{figure}

\begin{figure}[h]
    \centering
    \includegraphics[width=\textwidth]{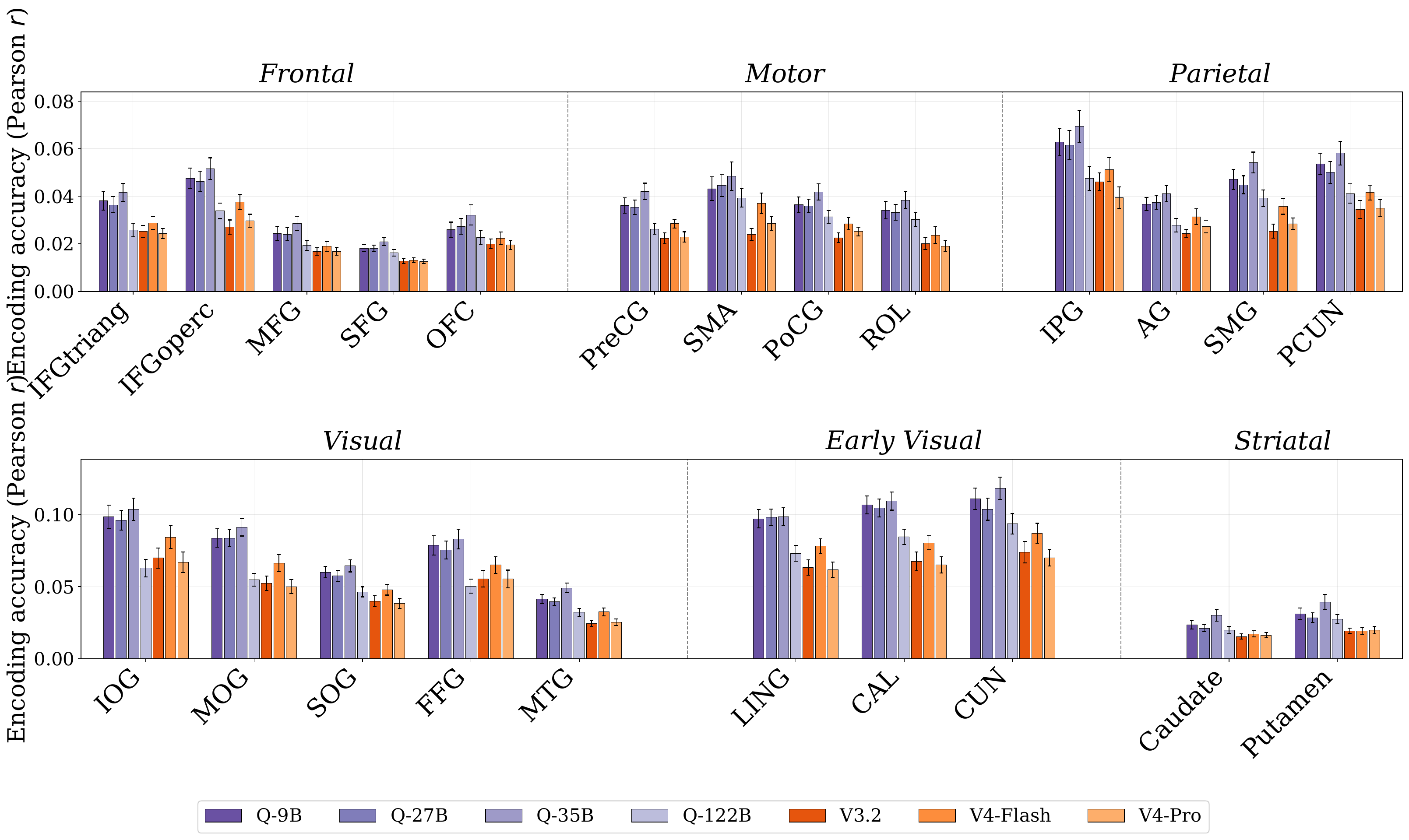}
    \caption{%
        \textbf{Per-ROI encoding accuracy for the seven LRMs.}
        Same conventions as Supp.~Fig.~\ref{supp:rois_baseline}, but
        with each bar showing the best-layer encoding accuracy for one
        of the seven LRMs. All LRMs achieve consistently higher
        encoding accuracy than the RL baselines across all ROIs, with
        the most pronounced effects in occipital and parietal regions
        (IOG, MOG, SOG, FFG, CAL, LING, CUN). The two DeepSeek thinking
        models V3.2 and V4-Flash systematically underperform the Qwen
        family and DeepSeek-V4-Pro across regions; we discuss possible
        reasons in the main text.%
    }
    \label{supp:rois_lrms}
\end{figure}

\begin{figure}[h]
    \centering
    \includegraphics[width=\textwidth]{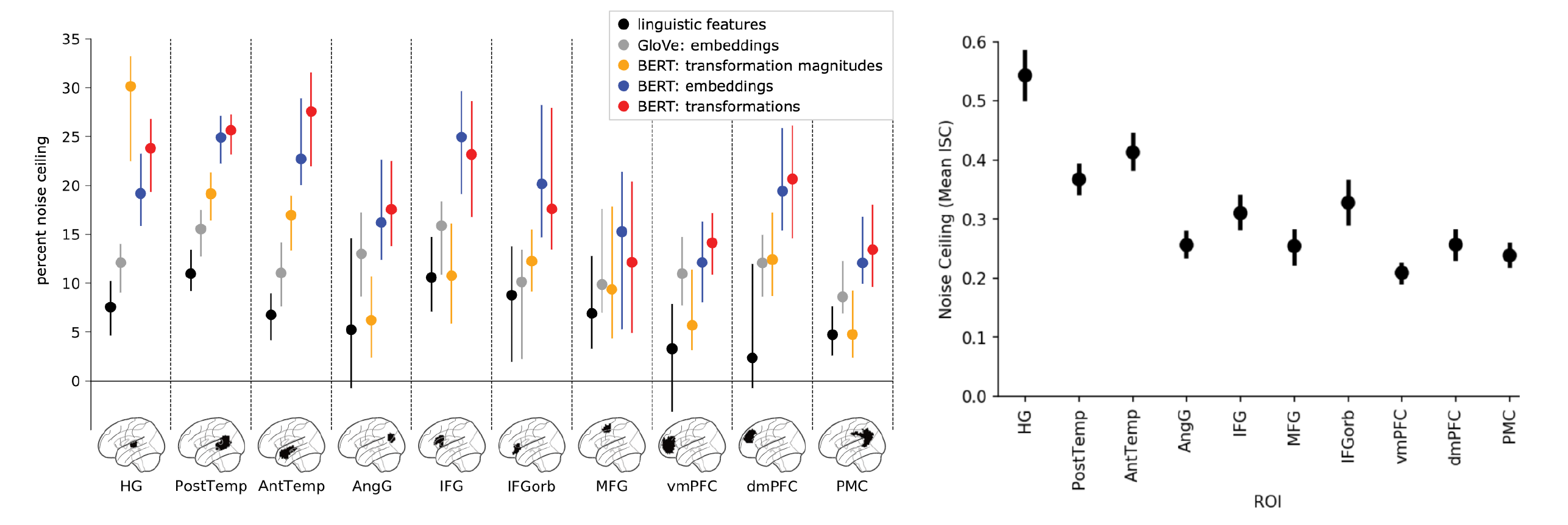}
    \caption{%
        \textbf{Comparison to prior fMRI encoding work using transformer
        representations during podcast story listening.}
        Figure reproduced from prior work by
        \citet{kumar2024shared}. Voxelwise banded ridge regression
        (same family of method as ours, reporting main band performance controlling for domain-specific
        nuisance regressors) using BERT embeddings (blue) and BERT
        attention-head transformations (red) to predict BOLD activity
        while subjects listened to spoken stories. \textbf{(Left)}
        Encoding accuracy expressed as percent noise ceiling
        ($\%\mathrm{NC}=r/r_{\mathrm{ceiling}}$) per ROI. \textbf{(Right)}
        Per-ROI noise ceiling, computed as the mean inter-subject
        correlation (ISC). Together these allow approximate recovery of
        absolute Pearson $r$ for the strongest BERT condition in each ROI:
        $r \approx \mathrm{ISC} \times (\%\mathrm{NC}/100)$.
        Comparing to our encoding accuracies in (Figure~\ref{supp:rois_lrms}) in shared or anatomically-similar
        ROIs: in MFG ($r \approx 0.04$ both papers) and angular gyrus
        ($r \approx 0.04$ both papers) the absolute values are nearly
        identical; in PCUN our LRMs achieve roughly twice the prior work's
        PMC value ($r \approx 0.06$ vs.\ $\approx 0.03$); in classical
        language regions our values are lower (IFG: $r \approx 0.04$ vs.\
        $\approx 0.08$; AntTemp/MTG: $r \approx 0.05$ vs.\ $\approx 0.12$),
        consistent with the difference in paradigm: \emph{language comprehension} vs.
        \emph{interactive game-learning}. The point of this comparison
        is not to claim equivalence between paradigms, but to establish
        that the absolute $r$ magnitudes reported in our main figures
        are within the range observed in published fMRI--language-model
        encoding work, rather than exceedingly low.%
    }
    \label{supp:podcast_bert_reproduction}
\end{figure}

\begin{figure}[h]
    \centering
    \includegraphics[width=\textwidth]{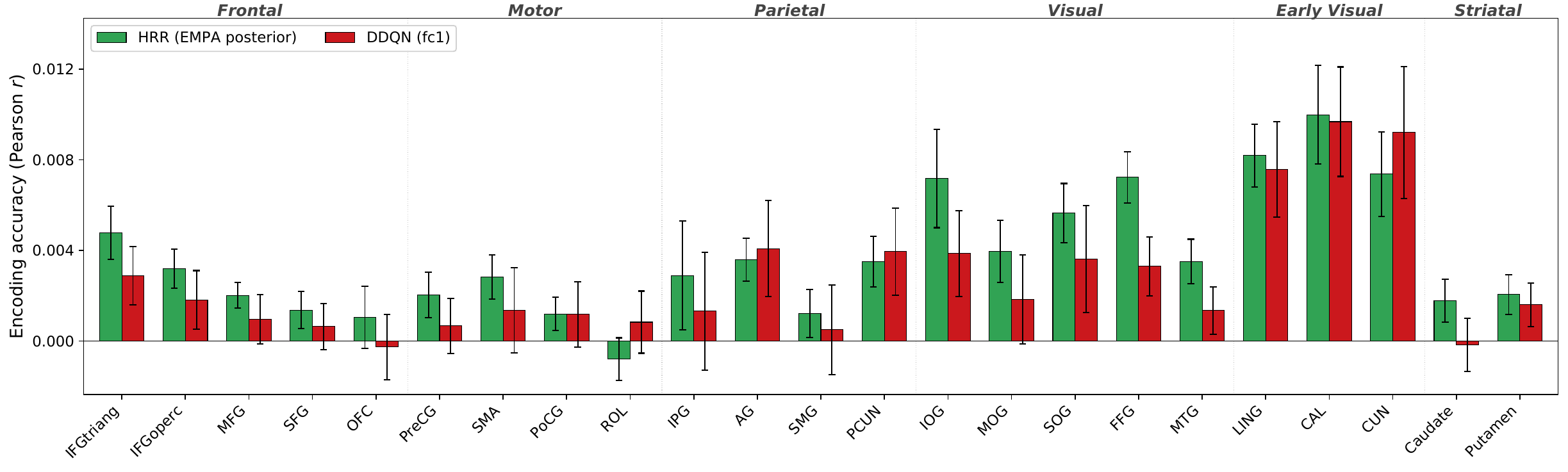}
    \caption{%
        \textbf{Reproducing the HRR > DDQN result of \citet{tomov2023neural}.}
        Per-ROI encoding accuracy (Pearson $r$, $\pm$ SEM across $n=21$) for the HRR baseline (green; Holographic Reduced Representation of EMPA's posterior over game
        theories) versus the fc1 layer of DDQN (red; same layer used
        in the original analysis). HRR's mean encoding accuracy exceeds
        DDQN's in 19 of 23 ROIs (one-sided sign test, $p<0.002$),
        consistent with the headline finding of \citet{tomov2023neural}
        that EMPA's theory-induction posterior carries more brain-aligned
        structure than DDQN's fc1 representation in "theory-coding regions" (IFG, FFG, etc). %
    }
    \label{supp:hrr_ddqn_repro}
\end{figure}
\begin{figure}[h]
\centering
\includegraphics[width=\textwidth]{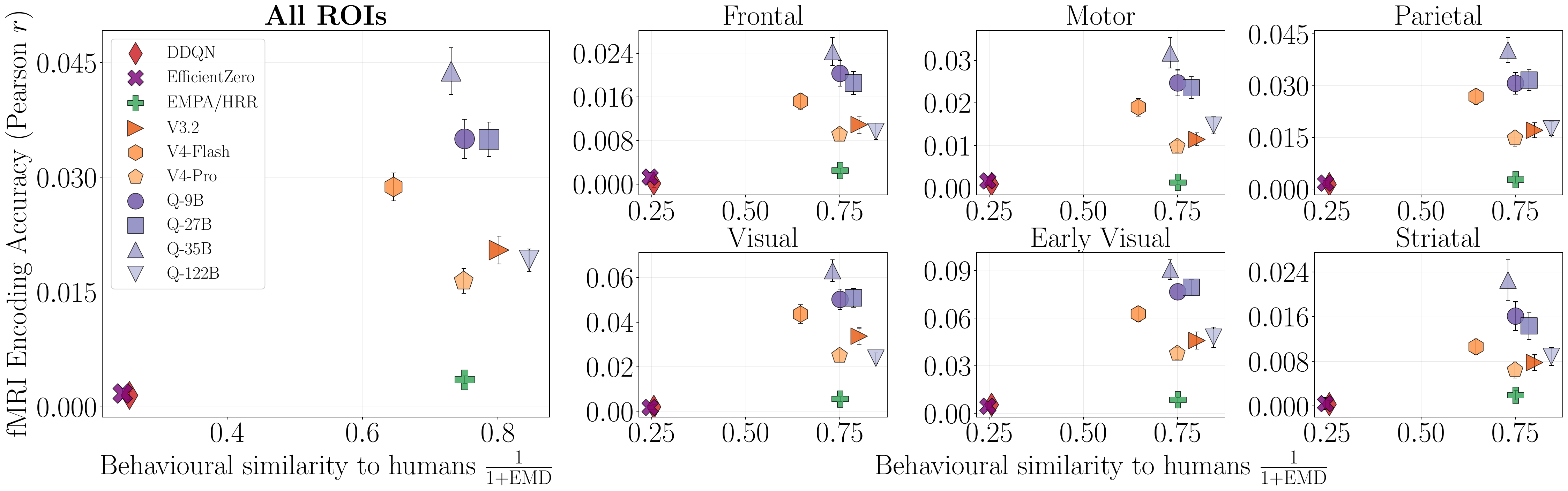}
\caption{%
  \textbf{Behavioural similarity vs.\ fMRI encoding accuracy.}
  Each point is one model; $x$-axis shows behavioural similarity to humans ($1/(1{+}\mathrm{EMD})$, log-space Wasserstein on discovery times, observed wins only);
  $y$-axis shows fMRI encoding accuracy (Pearson $r$, median layer per ROI, averaged across ROIs).
  Left panel aggregates across all ROIs; right panels break down by brain region.
  Error bars denote SEM across $n{=}21$ subjects.
  Models that play more like humans also tend to produce representations that better predict brain activity, with the strongest alignment in visual and early visual regions.
}
\label{fig:encoding_vs_behaviour}
\end{figure}

\clearpage

\subsection{Encoding control analyses}
\begin{figure}[h]
    \centering
    \includegraphics[width=\textwidth]{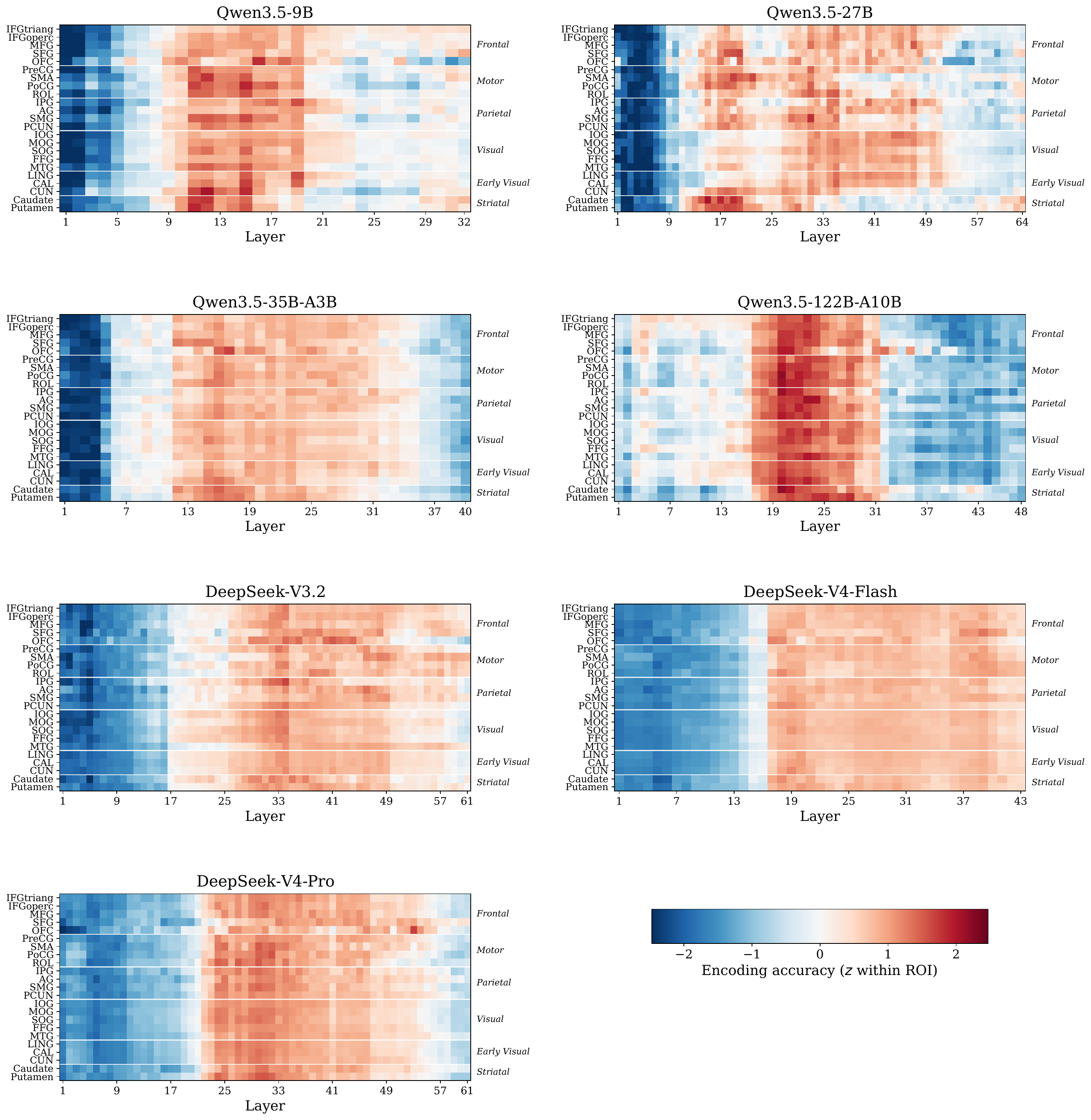}
    \caption{%
        \textbf{Layer~$\times$~ROI encoding heatmaps for the seven LRMs,
        $z$-scored within ROI.}
        For each LRM, mean encoding accuracy is computed at each
        (layer, ROI) cell (averaged across subjects and CV partitions,
        band~=~main), then $z$-scored within each ROI to remove the
        per-region absolute-magnitude differences (i.e.\ each row is
        normalised to zero mean and unit variance across that model's
        layers). Red cells thus indicate layers above the ROI's mean
        encoding accuracy and blue cells indicate layers below. This
        normalisation lets per-ROI layer patterns be read off
        independently of overall encoding strength: visual ROIs (IOG,
        MOG, SOG, FFG, CAL, LING, CUN) consistently peak at early-to-mid
        layers, while frontal and parietal ROIs (IFGtriang, MFG, SFG,
        IPG, AG, SMG, PCUN) peak at later layers. Subplot panels are
        sized by each model's native layer count (Qwen3.5-9B: 32 layers;
        Qwen3.5-27B: 64; Qwen3.5-35B-A3B: 40; Qwen3.5-122B-A10B: 48;
        DeepSeek-V3.2: 61; V4-Flash: 43; V4-Pro: 61). Region groups
        (Frontal, Motor, Parietal, Visual, Early Visual, Striatal) are
        labelled at right with thin separator lines.%
    }
    \label{supp:heatmaps}
\end{figure}

For all of these analyses, we evaluated on a subset of layers of each model corresponding to every $1/7$th depth, a sampling that makes it equitable to compare across models of different sizes and also hits the top performing layers of each of the models (See Figure~\ref{supp:heatmaps}). Layer subsets are matched between models and their control variants(Q-9B: layers
        \{1,5,11,16,21,27,32\}; Q-27B: \{1,11,21,32,43,53,64\};
        Q-35B-A3B: \{1,7,13,20,27,33,40\}).
\label{supp:controls}

\begin{figure}[h]
    \centering
    \includegraphics[width=\textwidth]{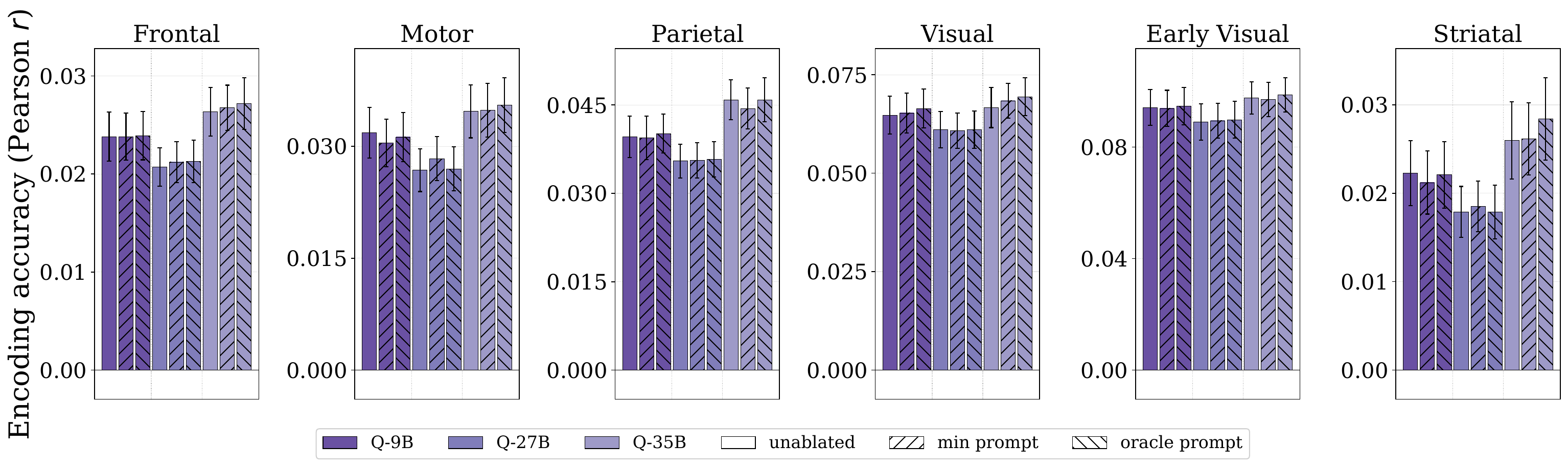}
    \caption{%
        \textbf{Encoding accuracy is robust to prompt-scaffolding choice.}
        For each of three Qwen families (9B, 27B, 35B-A3B), we compare
        encoding accuracy under the default (\texttt{full}) prompt
        scaffold against two ablations: minimal
        suggestion prompt and oracle suggestion
        prompt. Bars show group-mean best-layer encoding accuracy
        ($\pm$ SEM, $n=21$) per region group, with the parent model
        restricted to the layer set common across all three conditions
        in each triplet to avoid favoring it for having more extracted
        layers. Differences between prompt conditions are small
        relative to between-region differences, indicating that the
        encoding result is not an artefact of any specific prompt
        choice.%
    }
    \label{supp:prompt_ablation}
\end{figure}

\begin{figure}[h]
    \centering
    \includegraphics[width=\textwidth]{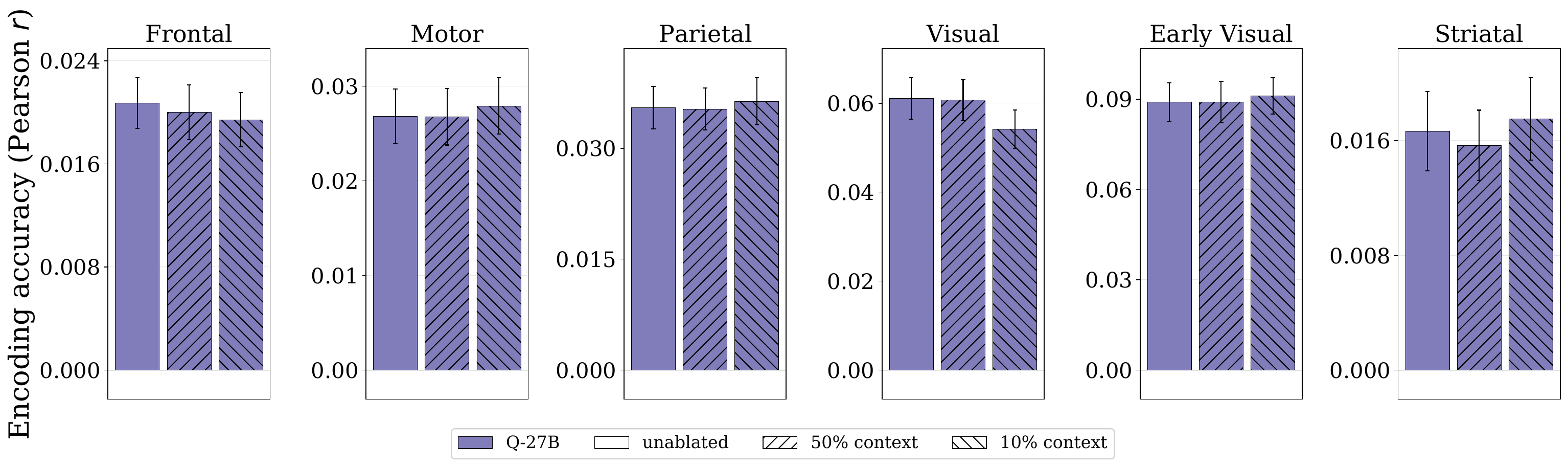}
    \caption{%
        \textbf{Encoding accuracy is robust to context-window length.}
        Qwen3.5-27B encoded with three context budgets: full context,
        50\% of full (\texttt{abl05}), and 10\% of full (\texttt{abl01}).
        Bars show group-mean best-layer encoding accuracy ($\pm$ SEM,
        $n=21$) per region group, with the parent model restricted to
        the common layer set across conditions. Encoding accuracy is
        approximately preserved at 50\% context and only modestly
        degrades at 10\% context, consistent with the encoding signal
        being driven by recent rather than long-range trajectory
        structure.%
    }
    \label{supp:context_ablation}
\end{figure}

\clearpage
\subsection{Layer-by-region structure and whole-brain surface maps}
\label{supp:heatmaps_surfaces}

\begin{figure}[h]
    \centering
    \includegraphics[width=\textwidth]{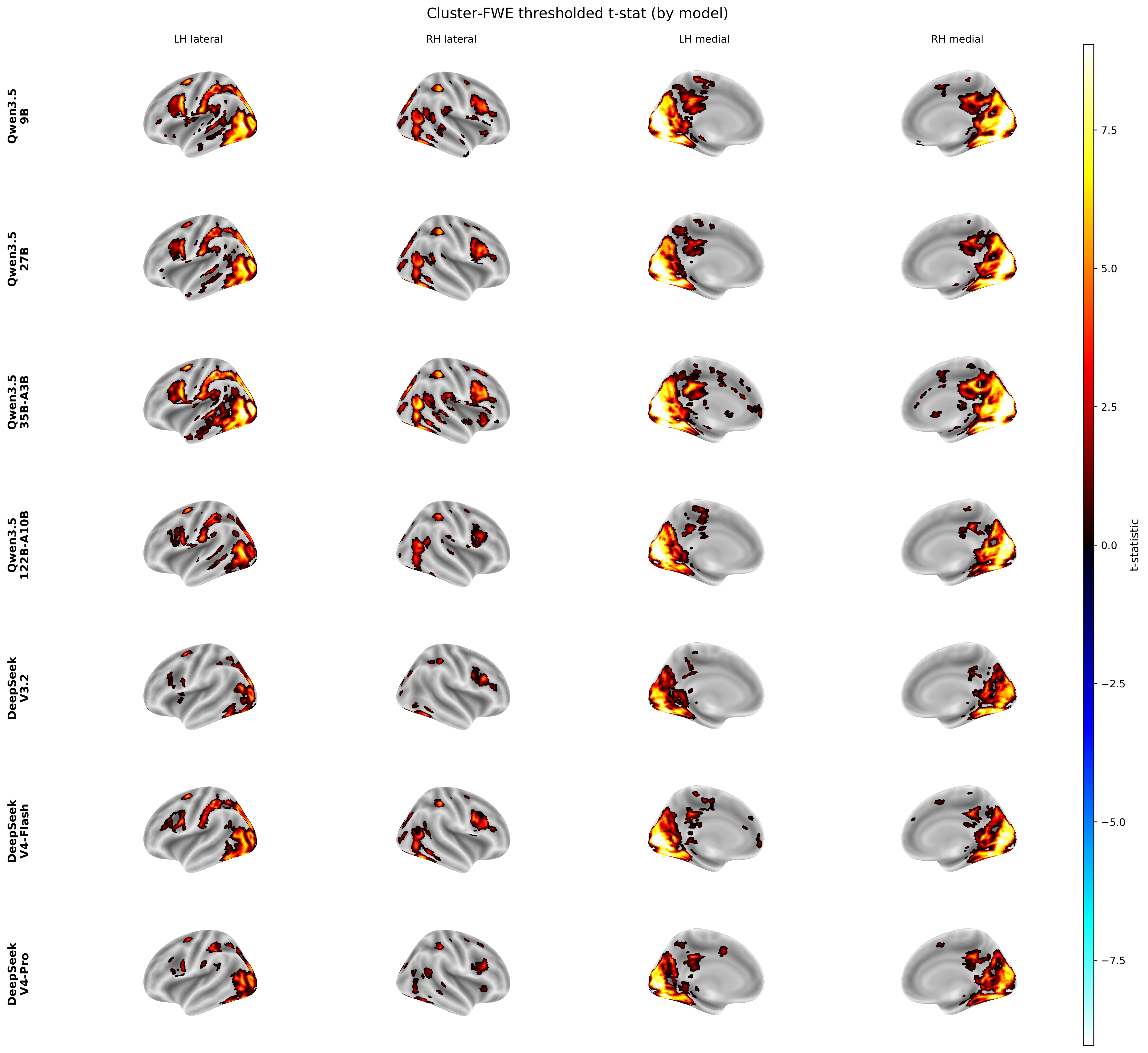}
    \caption{%
        \textbf{Cluster-FWE-thresholded $t$-statistic maps for all seven
        LRMs, rendered on the inflated \texttt{fsaverage} cortex.}
        Per-model voxelwise one-sample sign-flip permutation tests
        against zero ($n=21$ subjects, 5{,}000 sign-flip permutations,
        Fisher $z$-transformed, cluster-forming threshold $p<10^{-4}$,
        cluster-FWE corrected at $\alpha=0.05$). Each row is one LRM;
        the four columns show the standard views (LH-lateral,
        RH-lateral, LH-medial, RH-medial). The shared colour scale uses
        the 99\textsuperscript{th} percentile of $|t|$ pooled across
        all seven models, so visual contrast is directly comparable
        across rows. All seven models recover overlapping but
        non-identical sets of significant cortical territory: bilateral
        occipital cortex is consistently strongest, with secondary
        clusters in inferior frontal gyrus, supramarginal gyrus, and
        anterior cingulate cortex, and weaker but consistent involvement
        of premotor regions. Subcortical findings (striatum, thalamus,
        cerebellum) are not visible on cortical surface views; we
        report them anatomically in the main text.%
    }
    \label{supp:surface_all_llms}
\end{figure}

\begin{figure}[h]
    \centering
    \includegraphics[width=\textwidth]{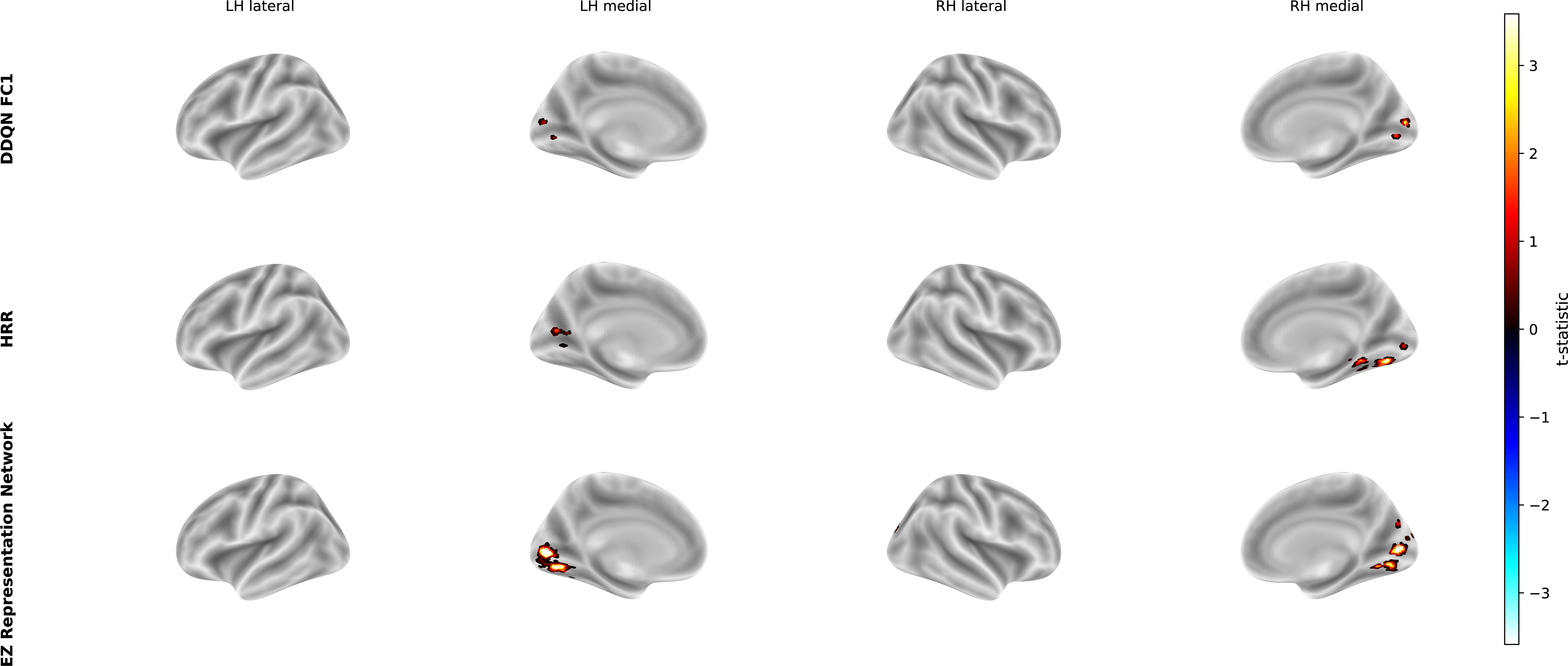}
    \caption{%
        \textbf{Cluster-FWE-thresholded $t$-statistic maps for Deep RL models.}
        Same as Figure~\ref{supp:surface_all_llms}, but for example DDQN and EfficientZero layers, as well as HRR. Since they don't survive the cluster-based multiple comparisons corrections, most voxels for these models do not have statistically significant prediction accuracy with the exception of small clusters in visual cortex. This is a stark contrast to the results reported in Figure~\ref{supp:surface_all_llms}.
    }
    \label{supp:surface_all_rl}
\end{figure}

\clearpage

\end{document}